%% file: main.tex
\documentclass[10pt,twocolumn,letterpaper]{article}

\usepackage[pagenumbers]{cvpr} 


\definecolor{cvprblue}{rgb}{0.21,0.49,0.74}
\usepackage[pagebackref,breaklinks,colorlinks,allcolors=cvprblue]{hyperref}
\usepackage{verbatim}
\usepackage{comment}
\usepackage{wrapfig}
\usepackage[accsupp]{axessibility}

\usepackage{graphicx}
\usepackage{amsmath}
\usepackage{amssymb}
\usepackage{booktabs}
\usepackage{bm}
\usepackage{adjustbox}
\usepackage{multirow}

\usepackage{url}            
\usepackage{amsfonts}       
\usepackage{nicefrac}       
\usepackage{microtype}      
\usepackage{xcolor}         

\usepackage{array}
\usepackage{xspace}

\usepackage[ruled]{algorithm2e} 

\input{macros.tex}

\def\paperID{6622} 
\def\confName{ICCV}
\def\confYear{2025}

\title{Human-Object Interaction from Human-Level Instructions}

\author{Zhen Wu \qquad Jiaman Li\qquad Pei Xu\qquad C. Karen Liu \\
Stanford University\\
{\tt\small \{zhenwu,jiamanli,peixu,karenliu\}@cs.stanford.edu}}

\begin{document}
\twocolumn[{%
\renewcommand\twocolumn[1][]{#1}%
\maketitle
\vspace{-12mm}
\begin{center}
    \centering
    \includegraphics[width=\textwidth]{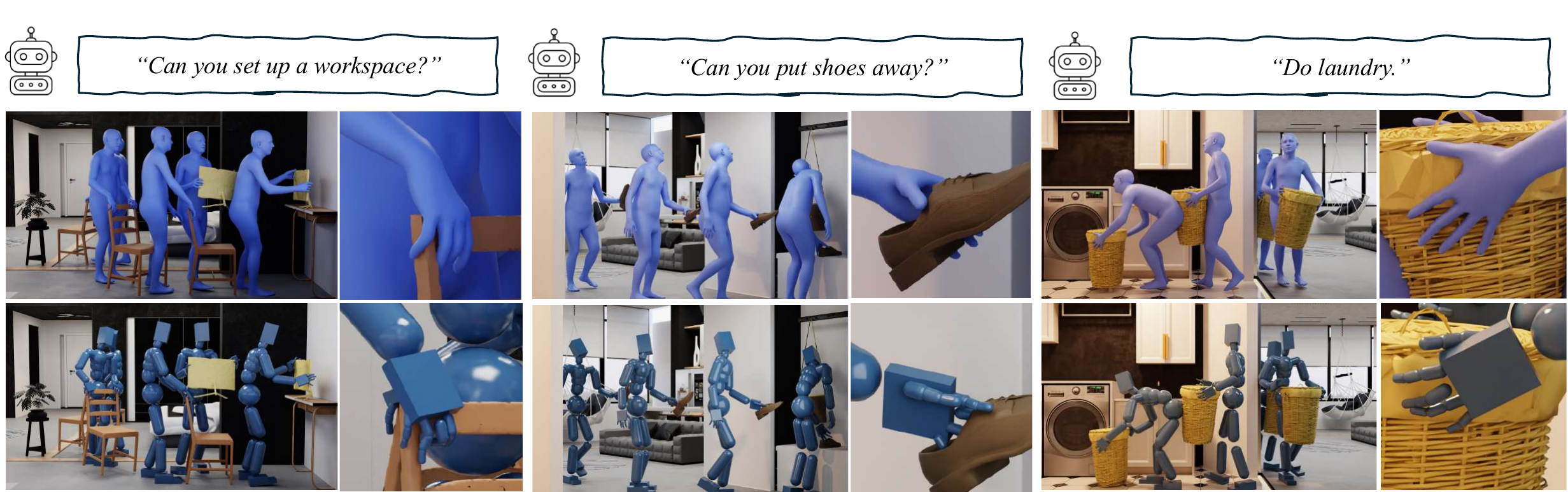}
    \vspace{-6mm}
    \captionof{figure}{Given human-level instructions (top), we first use a diffusion-based motion generator to synthesize natural full-body human motion, finger motion, and object motion to accomplish the task (middle). Next, we employ a physics-based tracking method to imitate the generated interactions which produces physically plausible interactions (bottom). Please refer to our \href{https://hoifhli.github.io/}{\textcolor{magenta}{project page}} for more results.}
    \label{figure1}
    \vspace{-6pt}
\end{center}%
}]

\setcounter{figure}{1}

\begin{abstract}
Intelligent agents must autonomously interact with the environments to perform daily tasks based on human-level instructions. They need a foundational understanding of the world to accurately interpret these instructions, along with precise low-level movement and interaction skills to execute the derived actions.
In this work, we propose the first complete system for synthesizing physically plausible, long-horizon human-object interactions for object manipulation in contextual environments, driven by human-level instructions. We leverage large language models (LLMs) to interpret the input instructions into detailed execution plans. Unlike prior work, our system is capable of generating detailed finger-object interactions, in seamless coordination with full-body movements. We also train a policy to track generated motions in physics simulation via reinforcement learning (RL) to ensure physical plausibility of the motion. Our experiments demonstrate the effectiveness of our system in synthesizing realistic interactions with diverse objects in complex environments, highlighting its potential for real-world applications.
\vspace{-2mm}

\end{abstract}

\input{chapters/1_introduction}
\input{chapters/2_related_work}

\input{chapters/3_system_overview}

\input{chapters/4_high_level_planner}
\input{chapters/5_low_level_interaction}

\input{chapters/7_low_level_finger}

\input{chapters/6_low_level_navigation}

\input{chapters/7_physics_tracker}
\input{chapters/8_experiment}

\input{chapters/9_conclusion}
\input{chapters/10_ack}

{
    \small
    \bibliographystyle{ieeenat_fullname}
    \bibliography{sample-bibliography}
}

\clearpage
\newpage
\input{supp}

\end{document}

%% file: macros.tex
\newcolumntype{L}[1]{>{\raggedright\let\newline\\\arraybackslash\hspace{0pt}}m{#1}}
\newcolumntype{C}[1]{>{\centering\let\newline\\\arraybackslash\hspace{0pt}}m{#1}}
\newcolumntype{R}[1]{>{\raggedleft\let\newline\\\arraybackslash\hspace{0pt}}m{#1}}

\newcommand{\ignore}[1]{}

\makeatletter
\DeclareRobustCommand\onedot{\futurelet\@let@token\@onedot}
\def\@onedot{\ifx\@let@token.\else.\null\fi\xspace}

\makeatother

\definecolor{MyDarkBlue}{rgb}{0,0.08,1}
\definecolor{MyDarkGreen}{rgb}{0.02,0.6,0.02}
\definecolor{MyDarkRed}{rgb}{0.8,0.02,0.02}
\definecolor{MyDarkOrange}{rgb}{0.40,0.2,0.02}
\definecolor{MyPurple}{RGB}{111,0,255}
\definecolor{MyRed}{rgb}{1.0,0.0,0.0}
\definecolor{MyGold}{rgb}{0.75,0.6,0.12}
\definecolor{MyDarkgray}{rgb}{0.66, 0.66, 0.66}

\definecolor{turquoise}{cmyk}{0.65,0,0.1,0.3}

%% file: chapters/1_introduction.tex
\section{Introduction}
\label{sec:introduction}

Synthesizing physically plausible, long-horizon, human-object interactions in contextual environments is crucial for applications in computer graphics, embodied AI, and robotics. When given human-level language instructions, intelligent agents should be able to comprehend these high-level commands and map them to executable action sequences. For instance, when instructed to set up a workspace, an agent must understand the abstract concept of ``workspace'', identify the relevant furniture and objects in the room, and translate this concept into a series of concrete actions, such as moving and orienting a desk, placing a chair nearby, and positioning a monitor on the desk. In addition to understanding the instructions, the agent must execute actions that lead to realistic and functional human movements, including navigating cluttered spaces, manipulating various objects for the task, and appropriately releasing them after use.

Prior research has made significant progress in synthesizing human-scene interactions using large language model (LLM) planners~\cite{xiao2024unified}. However, most of these interactions have been limited to static objects without hand manipulation. Some studies~\cite{li2023controllable} have demonstrated the manipulation of dynamic objects, but the resulting interactions still lack realistic finger movements. Recently, physics-based RL policies has demonstrated promising results in human-object interactions~\cite{luo2024grasping, wang2023physhoi}. However, generating long-horizon motions that interact with various objects in a scene-aware manner remains an ongoing challenge. Build upon these prior work, we introduce the first complete system capable of synthesizing physically accurate, long-horizon human-object interactions with synchronous full-body and intricate finger movements, from human-level language instructions.

Synthesizing coordinated full-body and finger movements from human-level instructions presents two main challenges. First, human-level instructions typically provide only a high-level task outline, requiring common sense and world knowledge to translate them into precise object arrangements and detailed execution plans. Second, existing datasets with paired full-body and finger motion data are limited to interactions with small objects and lack locomotion during object manipulation~\cite{GRAB:2020, fan2023arctic}.
Although a recent dataset~\cite{li2023object} includes locomotion and manipulation for large objects, it lacks detailed finger movements due to the difficulty of capturing both modalities simultaneously. 

To tackle the challenge of comprehending human-level instructions, we utilize the latest advancements in LLMs, which are highly effective in interpreting high-level commands\cite{openai2023gpt, liu2024visual}.
However, LLMs are not well-suited for directly predicting precise scene layouts, such as the exact positions and orientations of objects~\cite{aguina2024open, hong20233d}.
Instead of prompting LLMs to directly specify object placements, we use object spatial relationships as intermediate representations and ask LLMs to derive these relationships. We then propose an algorithm to calculate the precise object poses based on these relationships.

For the low-level motion synthesis, we take the approach of modeling human motion from real-world human data. To address the lack of large-scale object manipulation datasets with synchronized full-body and finger motions, one could use two separate datasets to train a full-body manipulation model and another detailed finger manipulation model. 
However, simply fusing the two motions together will cause glaring artifacts due to the disagreement in two separately trained models.
Our solution is to break this process into three steps.
We first generate full-body and object motions without detailed fingers. The object motion and rough wrist poses provide guidance for generating contact-consistent finger motion in the second step. Finally, we generate full-body and object motions again conditioned on precise finger motions. This approach enables our framework to successfully produce synchronized object motion, full-body motion, and finger motions without requiring a dataset that includes all data modalities.

The generated motion, however, is not guaranteed to be physically accurate.
To ensure physical plausibility, we further use reinforcement learning (RL) to track the generated motion in a physics simulator. 
Our method adopts an importance sampling technique to improve the training efficiency of long sequence tracking and supports tracking diverse interactions with multiple distinct objects within a single sequence.

We evaluate our method through extensive quantitative and qualitative comparisons, as well as human studies, demonstrating that our approach outperforms all others. This underscores the effectiveness of our method and its potential for real-world applications.



%% file: chapters/2_related_work.tex
\section{Related Work}
\label{sec:related_work}

\paragraph{Contextual Interaction Synthesis.}
Recent work on modeling human-object interactions falls into two categories: interactions with static and dynamic objects. Synthesizing human motions in static 3D scenes has been extensively studied~\cite{prox, wang2022humanise, araujo2023circle, jiang2024scaling, hassan_samp_2021, zhang2022couch}. 
Prior research has explored regression models~\cite{wang2021synthesizing, wang2022towards, hassan_samp_2021, araujo2023circle, mir2023generating} and diffusion models~\cite{huang2023diffusion, jiang2024scaling, yi2024generating, wang2024move} to generate human motions interacting with 3D scenes, such as sitting on a chair or lying down on a sofa. Some studies have also explored reinforcement learning methods to generate physically plausible human motions~\cite{hassan2023synthesizing, zhao2023synthesizing, xiao2024unified}. These works focus on static objects, whereas our work focuses on dynamic interactions.

There has been increasing attention on synthesizing human-object interactions with dynamic objects~\cite{diller2023cg, peng2023hoi, wu2024thor,xu2024interdreamer}. OMOMO~\cite{li2023object} collected a dataset for dynamic human-object interactions and proposed a framework to synthesize human motions from object motions. 
CHOIS~\cite{li2023controllable} synthesizes interactions in 3D scenes from text and sparse waypoints.
Despite these advancements, all of these works focus on full-body human motion generation without detailed finger motions. In this work, we aim to generate synchronized object motions, human motions, and finger motions.

\paragraph{Hand-Object Interaction Synthesis. }
Hand-object interaction synthesis has been extensively explored.
In the realm of static pose generation, traditional methods often rely on physics-based control or optimization techniques to generate grasp candidates \cite{li2007data, pollard2005physically, rodriguez2012caging}. 
Some recent work leverages deep learning to directly estimate hand grasps from extensive interaction datasets \cite{jiang2021hand, brahmbhatt2019contactgrasp, karunratanakul2020grasping, GRAB:2020, fan2023arctic, christen2024diffh2o}. 
Beyond static grasping, another line of work also explores dynamic object manipulation \cite{zhang2021manipnet, ye2012synthesis, liu2024geneoh}. 

Recent works further explore integrating whole-body motion and hand motion \cite{taheri2023grip, tendulkar2023flex, wu2022saga, taheri2022goal}.
IMoS \cite{ghosh2023imos} generates hand-object interaction using paired human motion and hand motion, while~\cite{braun2023physically} leverage reinforcement learning to achieve physically plausible motion.
However, these methods are constrained to manipulate small-sized objects in the GRAB dataset~\cite{GRAB:2020}, unable to generalize to large objects. 
In this work, we aim to generate realistic full-body and finger motions for manipulating large objects. Since no such paired dataset exists, we leverage DexGraspNet~\cite{wang2023dexgraspnet} to generate realistic grasp poses, train a conditional diffusion model on GRAB~\cite{GRAB:2020} to generate approaching and releasing finger motions, and then integrate these with our full-body motion generation.   

\paragraph{Physics-Based Motion Synthesis.}
Physics-based methods control a simulated character to perform motion in a simulator, ensuring the physical realism of the generated motion. To make the motion more natural, many approaches leverage motion capture data to produce movements that match the style of the data~\cite{peng2018deepmimic, peng2021amp, peng2023hoi, liu2018learning}. DeepMimic~\cite{peng2018deepmimic} proposes to track motion capture data using RL. Subsequent works~\cite{peng2021amp, xu2021gan} leverage generative adversarial networks (GANs) to learn motion styles. Recent research also explores this approach for human-object interactions, such as playing musical instruments~\cite{xu2024synchronize, wang2024f} or playing soccer~\cite{xie2022learning}.  PhysHOI~\cite{wang2023physhoi} learns basketball-playing skills from human demonstrations. OmniGrasp~\cite{luo2024grasping} enables a character to grasp and move an object along a specified object trajectory.
In this work, we leverage RL to track generated kinematic motion, successfully producing long-horizon motions that interact with various objects.

\paragraph{LLM-based Planning.}
LLMs have been widely used in reasoning, planning tasks~\cite{hong20233d, hu2023look, wake2023gpt, zeng2023distilling, ding2023task} for their high-level planning capabilities.
They have also been adopted in motion synthesis tasks~\cite{yao2024moconvq, zhang2024motiongpt, xiao2024unified, xu2024interdreamer}.
UniHSI~\cite{xiao2024unified} uses LLMs to extract action plans in a chain-of-contact format, guiding a low-level controller to synthesize corresponding motions. 
InterDreamer~\cite{xu2024interdreamer} employs LLMs to identify object parts that humans interact with and uses this information to retrieve initial human poses. 
In this work, we leverage LLMs to translate abstract human-level instructions into target scene maps and detailed execution plans, which guide the low-level interaction synthesis.

%% file: chapters/3_system_overview.tex
\section{System Overview}
\label{sec:system_overview}

\begin{figure*}[t]
    \centering
    \includegraphics[width=\textwidth]{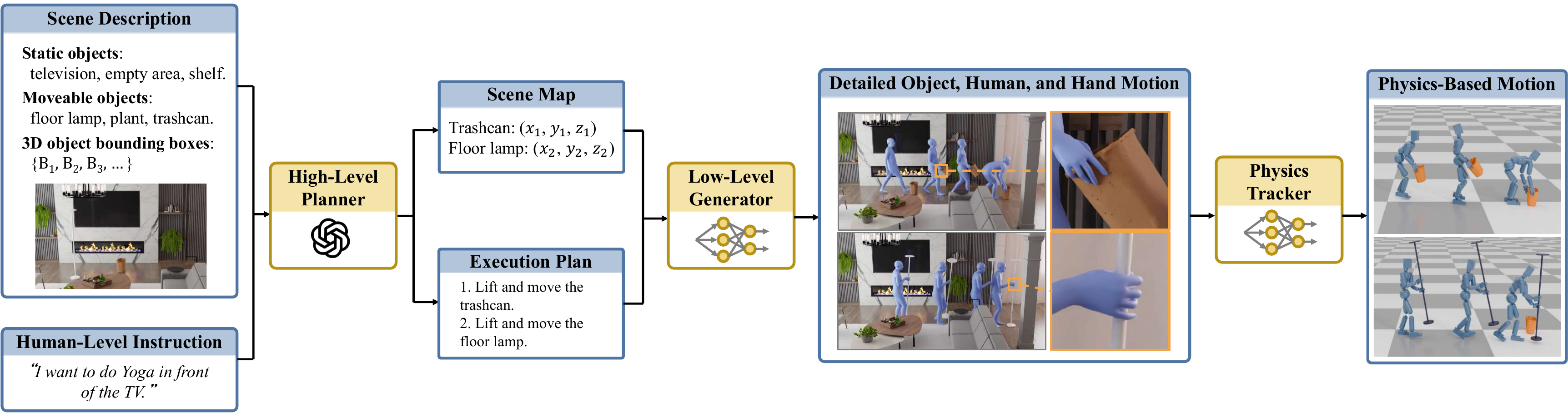}
    \vspace{-6mm}
    \caption{
    Our system takes the scene description and human-level instruction as input and uses a high-level planner to obtain the scene map and a detailed execution plan. The low-level motion generator then generates synchronized object motion, full-body human motion, and finger motion. Finally, the physics tracker uses RL to track the generated motion, producing physically plausible motion.}
    \vspace{-4mm}
    \label{fig:system_overview}
\end{figure*}
Given a human-level instruction and a scene description, we develop a system that generates synchronized human motion and object motion to accomplish the task. 
An overview of our system is depicted in \cref{fig:system_overview}. The system consists of a high-level planner, a low-level motion generator, and a physics tracker. The high-level planner leverages LLMs to reason about instructions and the scene description, generating a \emph{scene map} along with a detailed \emph{execution plan}. The scene map specifies the desired positions and orientations of the objects, while the execution plan outlines the steps to interact with these objects. The low-level motion generator unites conditional diffusion models trained on separate datasets for full body motion and dexterous hand motion, producing synchronous object and human movements.
The physics tracker then uses RL to track the generated motion in a physics simulator, ensuring physical plausibility.

%% file: chapters/4_high_level_planner.tex
\section{High-Level LLM Planner}
\label{sec:high_level_planner}

Inputs to the high-level planner consist of an initial scene description and human-level instructions. The scene description includes a list of objects in a 3D scene and their layout, while the human-level instruction describes a task that requires grounded reasoning based on world knowledge and common sense. For example, the instruction ``I want to do yoga in front of the TV'' implies that any obstacles in front of the TV should be moved elsewhere.

Outputs of the high-level planner comprise the scene map and the execution plan.
The scene map specifies the target position $p$ and orientation $q$ for each object $o$ that requires movement, represented as $\left\{ \left\{o_1, p_1, q_1 \right\}, \cdots, \left\{o_n, p_n, q_n \right\}\right\}$.
The execution plan is a sequence of text actions associated with objects in the scene map. For example, “lift the monitor, move the monitor, put down the monitor”.

\subsection{Scene Map Generation}
\label{subsec:spatial_relationship}

To obtain a reliable scene map, inspired by recent works in room layout generation \cite{aguina2024open, yang2024holodeck}, we first instruct LLMs to reason about spatial relationships of the objects as an intermediate representation from which the 3D positions and orientations of the objects are calculated subsequently.
We define the following relation functions to capture spatial relationships, with flexibility to include additional relationships if needed.

\begin{enumerate}
    \item \texttt{on(object1, object2)}.
    \item \texttt{adjacent(object1, object2, direction, distance)}.
    \item \texttt{facing(object1, object2)}.
\end{enumerate}

For example, if a monitor needs to be put on the table facing the chair, we can use \texttt{on(monitor, table)} and \texttt{facing(monitor, chair)} to describe its pose. If a table is positioned 1 meter to the north of the door, we can use \texttt{adjacent(table, door, north, 1)} to describe the relationship. 
We prompt LLMs to generate the spatial relationships of the objects according to the given instruction. 


Given the spatial relationships, we propose an algorithm to calculate the 3D positions of each object in the scene. The orientations of objects, enforced by ``facing'' relations, will be addressed in the subsequent step. 

The algorithm first constructs a scene graph \cite{johnson2015image} to organize the objects (nodes) and their spatial relationships (edges), which are directional, pointing from \texttt{object2} to \texttt{object1}. The algorithm employs a method similar to topological sorting, assuming no cycles are present in the graph. The scene graph is initialized by visiting the nodes associated with static objects and setting their 3D positions as the given values. In each iteration, the algorithm looks for a new node whose 3D position has not been determined but all its predecessors have. By calculating an offset from the predecessors' known positions, we can determine the position of the new node. For example, if object $o_1$ is ``on'' object $o_2$, then $o_1$'s height is set to match the top surface of $o_2$, while the horizontal position is sampled within the horizontal range of $o_2$. If $o_1$ is ``adjacent'' to $o_2$, we use the direction and distance given by the LLMs as the offset. We have included pseudo-code in the supplementary material.

To compute the orientation enforced by \texttt{facing($o_1$, $o_2$)}, we rotate the canonical direction of $o_1$ in its own local frame to align with the vector from the position of $o_1$ to the position of $o_2$. The canonical direction of each object is part of the input 3D geometry information.

\subsection{Execution Plan Generation}
\label{subsec:plan}
 
After determining the scene map, the planner must specify the interaction order with each object to ensure natural motion. For instance, if a vase is on a table, the vase should be moved before the table. 
LLMs are asked to generate this action sequence in a natural order and provide reasoning for it.
The resulting execution plan consists of textual actions, $\left\{l_1, l_2, \dots, l_{T}\right\}$, where each action $l_t$ defines interaction involving one object. 
Each $l_t$ serves as the textual input for the low-level generator, following the format: ``lift the \textit{object}, move the \textit{object}, put down the \textit{object}''. This format is designed to closely match the textual style of the training dataset of the low-level generator.

With the scene map and execution plan in place, an A* planner generates collision-free paths to execute the plan. The path is a series of 2D waypoints in the scene, which then serve as inputs for the low-level motion generator.

%% file: chapters/5_low_level_interaction.tex
\section{Low-Level Motion Generator}

\begin{figure*}[t]
    \centering
\includegraphics[width=\textwidth]{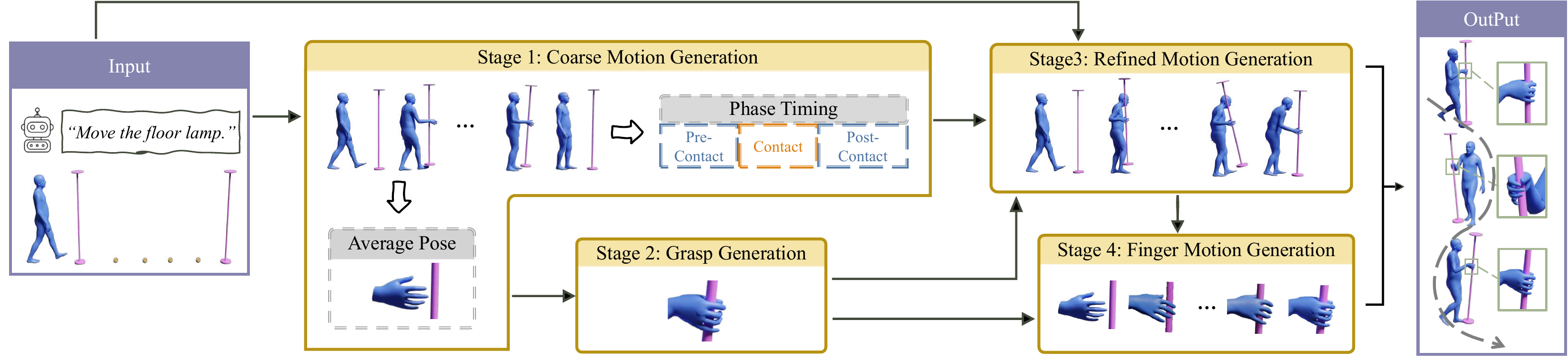}
\vspace{-6mm}
    \caption{The interaction module consists of four stages. Starting with the initial human and object poses, waypoints, final object pose, and a text instruction, the first stage generates coarse object and human motion. The relative wrist pose from this motion is then used to generate a valid grasp pose. Using this as a condition, the next stage re-generates the motion to align with the grasp constraint. Finally, the finger motion generation module produces finger motions during the approach and release of objects.} 
    \vspace{-4mm}
    \label{fig:coarse_to_fine}
\end{figure*}

Given the waypoints and textual instructions generated by the high-level planner, the low-level motion generator executes an interaction module and a navigation module sequentially to generate long-horizon human-object interaction motions.

\subsection{Background: Conditional Diffusion Model}
\label{sec:cond_diff}
The diffusion model involves a forward diffusion process and a reverse diffusion process. Starting from clean data $\bm{x}_0$,  noise is progressively introduced until, after N steps, the data becomes approximately $\bm{x}_{N} \sim \mathcal{N}(0, \bm{I})$. The denoising network is trained to reverse this process so that sampling a noise $\bm{x}_{N}$ and applying N denoising steps yields the clean data representation $\hat{\bm{x}}_0$. Each denoising step is defined as 
\begin{equation}
p_{\theta}(\bm{x}_{n-1}|\bm{x}_{n}, \bm{c}) := \mathcal{N}(\bm{x}_{n-1}; \bm{\mu}_{\theta}(\bm{x}_n, n, \bm{c}), \sigma_{n}^{2}\bm{I}),  
\end{equation}
where $p_{\theta}$ represents a neural network, $\bm{c}$ represents the conditions, $\bm{\mu}_{\theta}(\bm{x}_n, n, \bm{c})$ represents the learned mean, $\sigma_{n}^{2}$ denotes a fixed variance. The training loss is defined as $\mathcal{L} =  \mathbb{E}_{\bm{x}_0, n}||\hat{\bm{x}}_{\theta}(\bm{x}_{n}, n, \bm{c}) - \bm{x}_{0}||_{1}$.

\subsection{Interaction Module}
\label{sec:low_level_interaction}
We introduce a multi-stage interaction module that synthesizes synchronized full-body, finger, and object motion, as shown in~\cref{fig:coarse_to_fine}.
The first stage, \textit{CoarseNet}, generates initial human and object motions without detailed finger movement.
Recognizing that finger poses typically remain static during interaction, the second stage generates a \textit{grasp} pose based on the initial results.
Then \textit{RefineNet} re-generates the human and object motions to seamlessly align with the optimized grasp pose.
Finally, \textit{FingerNet} generates smooth finger motions for approaching and releasing the object, completing the interaction sequence.

\subsubsection{CoarseNet}
\label{sec:coarse_net}
Given the initial human and object poses, waypoints, and final object pose, along with a text instruction, CoarseNet generates synchronized human motion and object motion. This is achieved using a pre-trained diffusion model CHOIS~\cite{li2023controllable}.

\paragraph{Data Representation.} 
We denote human motion as $\bm{H} \in \mathbb{R}^{T \times D}$, where $T$ represents the sequence length and $D$ represents the dimension of the human pose. Each $\bm{H}_{t}$ includes global joint positions and 6D rotations~\cite{zhou2019continuity}, excluding finger joints. Object motion $\bm{O} \in \mathbb{R}^{T \times 12}$ includes the object's global position and rotation matrix.
CoarseNet predicts human motion, object motion, and contact labels $\bm{L} \in \mathbb{R}^2$ for both hands, resulting in $\bm{x} = \left\{\bm{H}, \bm{O}, \bm{L}\right\}$.

The input condition is represented as $\bm{c} = \left\{\bm{S}, \bm{G}, \bm{T}\right\}$. $\bm{S} \in \mathbb{R}^{T \times (D+12)}$ is a masked motion data representation. It includes the human and object pose in the first frame, the object pose in the last frame, and 2D waypoints sampled every 30 frames. Any remaining entries in $\bm{S}$ are padded with zeros. The object geometry $\bm{G}$ represents the BPS representation~\cite{prokudin2019efficient} to encode object geometry, and $\bm{T}$ refers to the text embeddings extracted using CLIP~\cite{radford2021learning}.

\vspace{-3mm}
\paragraph{Segmenting Contact Phases.} 
We segment the generated interaction sequence into \textit{pre-contact}, \textit{contact}, and \textit{post-contact} phases based on the predicted contact labels $\bm{L}$.  
The contact phase is determined separately for each hand, dividing the model output $\bm{x}_\text{coarse}$ into $\bm{x}_\text{pre-contact} \oplus \bm{x}_\text{contact} \oplus \bm{x}_\text{post-contact}$.
We then compute the average wrist pose $\bm{w}$ from $\bm{x}_\text{contact}$, serving as input for the next stage. 

\subsubsection{Grasp Pose Generation}
\label{subsubsec:grasp_pose}

The second stage generates the finger pose for the contact phase using a state-of-the-art grasp pose generation method~\cite{wang2023dexgraspnet}.
Given an object, wrist pose $\bm{w}$ and rest finger pose, an optimization is conducted to generate a physically plausible grasp pose.  
We modified \cite{wang2023dexgraspnet} to fit our needs by: (1) increasing the object surface sample points from 2,000 to 20,000 to maintain sample density for larger objects and (2) omitting the force closure term for tasks involving both hands, such as carrying a box, where stability doesn’t solely rely on a single hand.

The optimized grasp pose is denoted by $\hat{\bm{g}} = (\hat{\bm{w}}, {\hat{\bm{\theta}}})$, where $\hat{\bm{w}}$ is the optimized wrist pose, and $\hat{\bm{\theta}}$ is the finger pose. This grasp pose $\hat{\bm{g}}$ is maintained throughout the contact phase. However, as CoarseNet’s generated motion may not fully align with $\hat{\bm{w}}$, we introduce RefineNet to address this misalignment.

\subsubsection{RefineNet}
\label{sec:refine_net}


The goal of RefineNet is to re-generate human-object interaction motions that align with the optimized grasp pose $\hat{\bm{g}}$. We adopt a conditional diffusion model based on CoarseNet, adding two conditions: \textit{wrist-object relative pose} to align the wrist with $\hat{\bm{w}}$ and \textit{object static} to keep the object stationary during non-contact phases. RefineNet’s input condition, $\bm{c}_{r} = \left\{\bm{W}, \bm{S}_{r}, \bm{G}, \bm{T}\right\}$, includes $\bm{G}$ and $\bm{T}$ from CoarseNet, with $\bm{W}$ and $\bm{S}_{r}$ detailed as follows.

\vspace{-3mm}
\paragraph{Wrist-Object Relative Pose Condition.} 
The condition $\bm{W} \in \mathbb{R}^{T \times 18}$ represents the position and 6D rotation of both wrists in the object's frame, including only contact-phase wrist poses, with other entries set to zero.

\vspace{-2mm}
\paragraph{Object Static Condition.}
In CoarseNet, the masked motion condition $\bm{S}$ was zero-padded except at the start, end, and waypoints. RefineNet uses contact phase information from CoarseNet to adjust $\bm{S}$ to $\bm{S_{r}}$, assigning a static object pose in pre-contact and post-contact frames.


\vspace{-2mm}
\paragraph{Wrist-Object Relative Pose Loss.}
We incorporate an additional loss to enforce the wrist-object relative pose condition. We sample 100 points $\bm{K}_{\text{rest}} \in \mathbb{R}^{100 \times 3}$ uniformly on the rest object’s surface and pre-calculate their positions in both wrists' coordinate frames, denoted as $\bm{K}_{\text{w}} \in \mathbb{R}^{2\times T \times 100 \times 3}$. 
At each time step, we compute the points' positions in the wrist frame, $\hat{\bm{K}}_{\text{w}}$, using the predicted orientations and positions of the object and wrist, $\bm{R}_{\text{o}}, \bm{T}_{\text{o}}, \bm{R}_{\text{w}}, \bm{T}_{\text{w}}$:
\begin{align}
\bm{K}_{\text{global}} &= \bm{R}_{\text{o}}\bm{K}_{\text{rest}} + \bm{T}_{\text{o}}, \\
\hat{\bm{K}}_{\text{w}} &= \bm{R}_{\text{w}}^{-1} \left( \bm{K}_{\text{global}} - \bm{T}_{\text{w}}\right).
\end{align}
The loss is computed as:
\begin{align}
\mathcal{L}_{\text{relative }}=
\sum_{t=1}^T
\bm{L}_{t}
\left\| 
\hat{\bm{K}}_{\text{w}, t}-\bm{K}_{\text{w}, t}
\right\|_1,
\end{align}
where $\bm{L}_{t}$ represents the contact labels, masking out the loss when the hand and object are not in contact.


\vspace{-2mm}
\paragraph{Post-Processing.}
RefineNet generates human-object motions that align closely with input conditions, though minor adjustments may be required. To refine alignment, we apply post-processing: replacing object poses in pre- and post-contact phases with static poses and recalculating wrist trajectories during the contact phase based on object motion and the optimized wrist pose $\hat{\bm{w}}$. Additional implementation details are in the supplemental materials.

%% file: chapters/7_low_level_finger.tex
\subsubsection{FingerNet}
\label{sec:low_level_finger}
In this stage, we generate finger motions for the pre-contact and post-contact phases to create natural transitions between the rest and grasp poses, forming a complete interaction sequence along with predictions in prior stages.
Given the start and end finger poses and the wrist trajectory in between, we employ a conditional diffusion model to predict the finger motions denoted as $\bm{F} \in \mathbb{R}^{T\times D'}$, representing local 6D rotations~\cite{zhou2019continuity}.




The model is conditioned on $\bm{c}_{f} = \left\{\bm{P}, \bm{F}_{s}, \bm{F}_{e}\right\}$, where $\bm{P}$ captures spatial hand-object relationships and $\bm{F}_{s}, \bm{F}_{e}$ are the start and end finger poses. Following prior work~\cite{zhang2021manipnet,taheri2023grip}, $\bm{P} \in \mathbb{R}^{T \times 100}$ is computed by sampling 100 palm-side mesh vertices and measuring their closest distances to the object, with the finger pose set to mean pose when calculating $\bm{P}$. 




%% file: chapters/6_low_level_navigation.tex
\subsection{Navigation Module}
\label{sec:low_level_navigation}
To generate long sequences of interactions with multiple objects, the human needs to navigate through the scene based on waypoints.
We adopt a conditional diffusion model to generate human navigation motion $\bm{H} \in \mathbb{R}^{T \times D}$. The input conditions consist of the initial human pose, 2D waypoints, and waypoint orientations (expressed as normalized direction vectors) to maintain consistent facing directions. 
Smooth transitions between the interaction and navigation modules are achieved by using the final pose of one module as the starting pose for the next.

%% file: chapters/7_physics_tracker.tex
\section{Physics Tracker}
After obtaining a sequence of kinematic motions, we train a tracking policy using reinforcement learning to control a physically simulated character that imitates these motions. This approach helps eliminate artifacts (e.g., hand-object penetration and foot sliding) in the kinematic motions and ensures the physical plausibility of the generated results.

We use PPO~\cite{schulman2017proximal} as the backbone reinforcement learning algorithm,
and adopt an importance sampling strategy to facilitate the tracking of long motion sequences involving interactions with multiple objects.
Policies trained by our approach can accurately track the kinematics motions, enabling the character to grasp and move distinct objects along the generated trajectories.
Implementation and training details for the physics tracker are provided in the supplementary materials.

%% file: chapters/8_experiment.tex
\section{Experiment}
\label{sec:experiment}
We conduct extensive evaluation to assess the overall performance of our system, as well as each individual component. We use quantitative metrics and user perception studies to evaluate our results compared to the baselines. We also conduct ablation studies to analyze the effects of each system module. 
We present the key results in this section and highly recommend readers refer to the supplementary video and documents for more detailed results and descriptions of the studies.

\subsection{Evaluation of High-Level Planner}
\label{subsec:planner_eval}



The core component of the high-level planner is the scene map calculation. To assess its effectiveness, we quantitatively measure the geometric accuracy of the resulting plans and conduct a human perception study.
For evaluation, we compare our method to a baseline that uses LLMs to directly predict object positions and orientations. We design 25 human-level instructions covering common everyday tasks for assessment, using OpenAI's GPT-4o \cite{gpt4o} for both our method and the baseline.

\begin{figure}[t]
\begin{center}
    \includegraphics[width=\columnwidth]{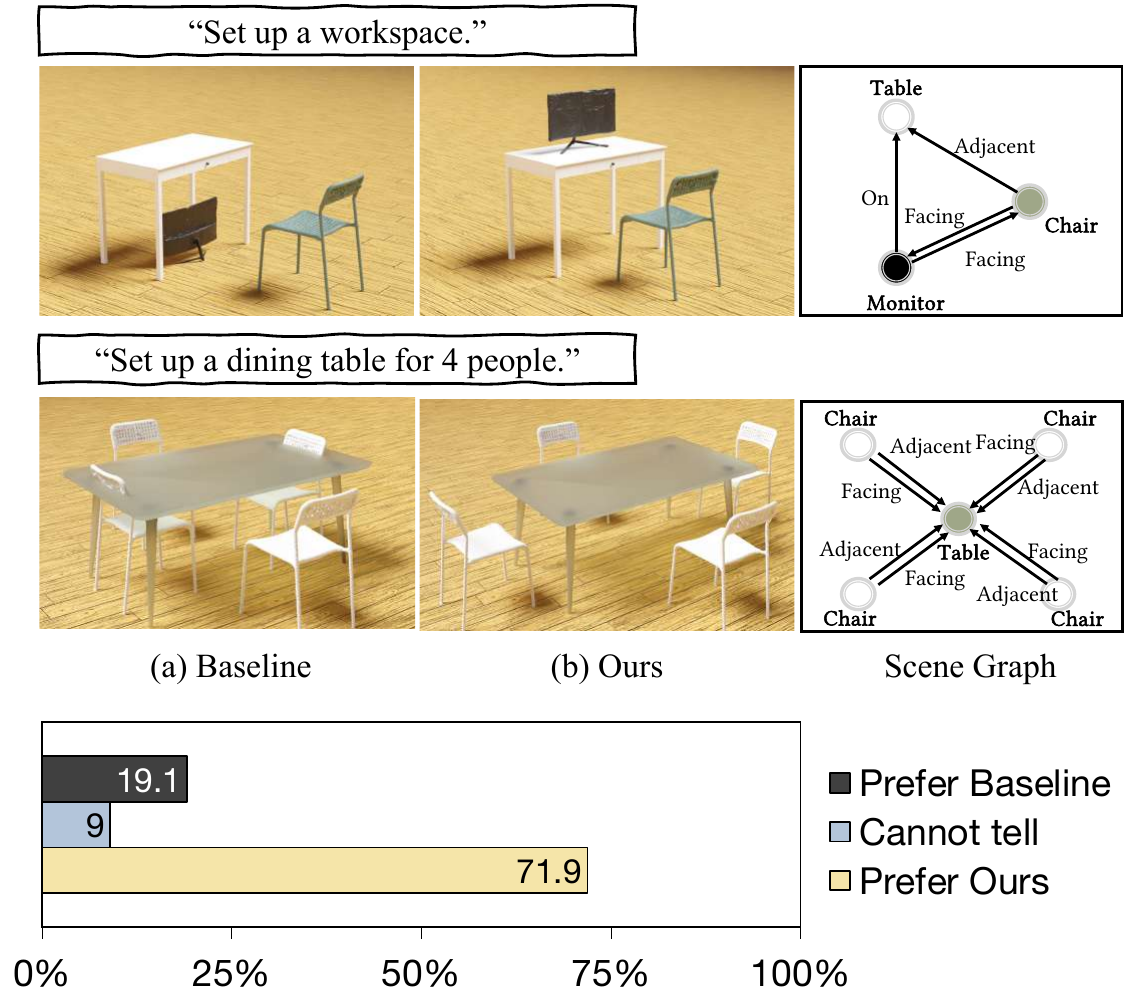}
    \vspace{-5mm}
    \caption{Comparison between (a) the baseline and (b) our high-level planner. The baseline generates an incorrect monitor position and orientation, and places the chair in a position that intersects with the table. The human perception study results (bottom) show that the majority of participants preferred our results.
    }
    \label{fig:llm_result}
   \vspace{-8mm}
\end{center}
\end{figure}

\begin{table*}[ht]
\centering
\begin{adjustbox}{max width=\textwidth}
\begin{tabular}{lcccccccccccccccccc}
\toprule
\multirow{2}{*}{Method} & \multicolumn{1}{c}{Condition} & \multicolumn{2}{c}{Human Motion} & \multicolumn{5}{c}{Interaction} & \multicolumn{3}{c}{GT Difference} \\
\cmidrule(r){2-2} \cmidrule(r){3-4} \cmidrule(r){5-9} \cmidrule(r){10-12}
& $T_{xy} \downarrow$ & $H_{\text{feet}} \downarrow$ & FS $\downarrow$ & $\text{C}_{\text{prec}} \uparrow$ & $\text{C}_{\text{rec}} \uparrow$ & $\text{C}_{\text{F1}} \uparrow$ & CC $\uparrow$ & IV $\downarrow$ & $\text{MPJPE} \downarrow$ & $T_{\text{obj}} \downarrow$ & $O_{\text{obj}} \downarrow$ \\
\midrule
CNet+GRIP \cite{taheri2023grip} 
& 3.58 & \textbf{2.50} & 0.45 & 0.90 & 0.64 & 0.70 & 3.99\% & 49.00 & 14.33 & 11.55 & 0.93 \\
\midrule
CNet 
& 3.26 & 2.68 & 0.43 & 0.91 & 0.81 & 0.84 & 3.54\% & 48.56 & \textbf{14.32} & \textbf{10.56} & 1.06 \\
C+RNet 
& \textbf{2.81} & 2.89 & \textbf{0.33} & \textbf{0.91} & \textbf{0.95} & \textbf{0.92} & 3.84\% & 24.78 & 15.96 & 12.57 & \textbf{0.73} \\
C+R+FNet (ours)
& \textbf{2.81} & 2.89 & \textbf{0.33} & \textbf{0.91} & \textbf{0.95} & \textbf{0.92} & \textbf{5.53\%} & \textbf{19.06} & 15.96 & 12.57 & \textbf{0.73}\\
\bottomrule
\end{tabular}
\end{adjustbox}
\caption{Interation synthesis on the FullBodyManipulation dataset \cite{li2023object}. Our full model achieves superior interaction quality compared to all others. }
\label{tab:interaction}
\end{table*}

\begin{figure*}[t]
    \centering
\vspace{-2mm}
\includegraphics[width=\textwidth]{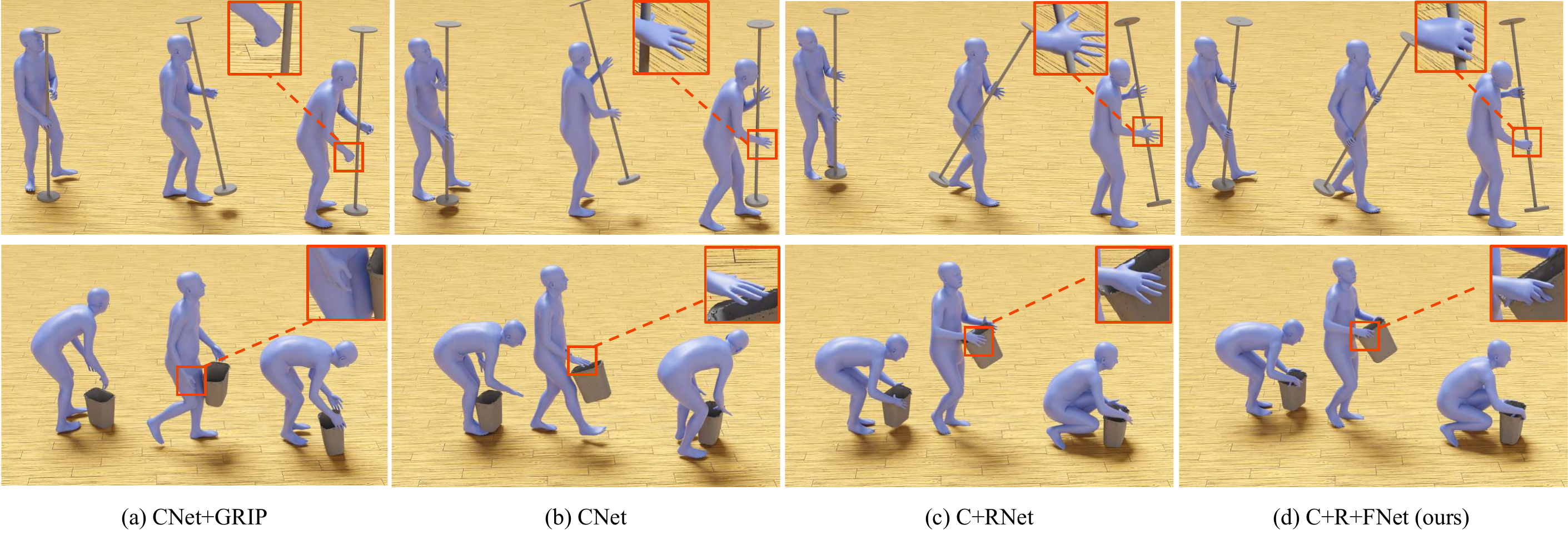}
\vspace{-8mm}
    \caption{Qualitative comparisons of each method. All baselines and ablations fail to produce precise hand and finger motions, while our method generates natural hand-object interactions.} 
    \label{fig:interaction_qualitative}
\vspace{-4mm}
\end{figure*}

\begin{figure}[t]
    \centering
\includegraphics[width=0.48\textwidth]{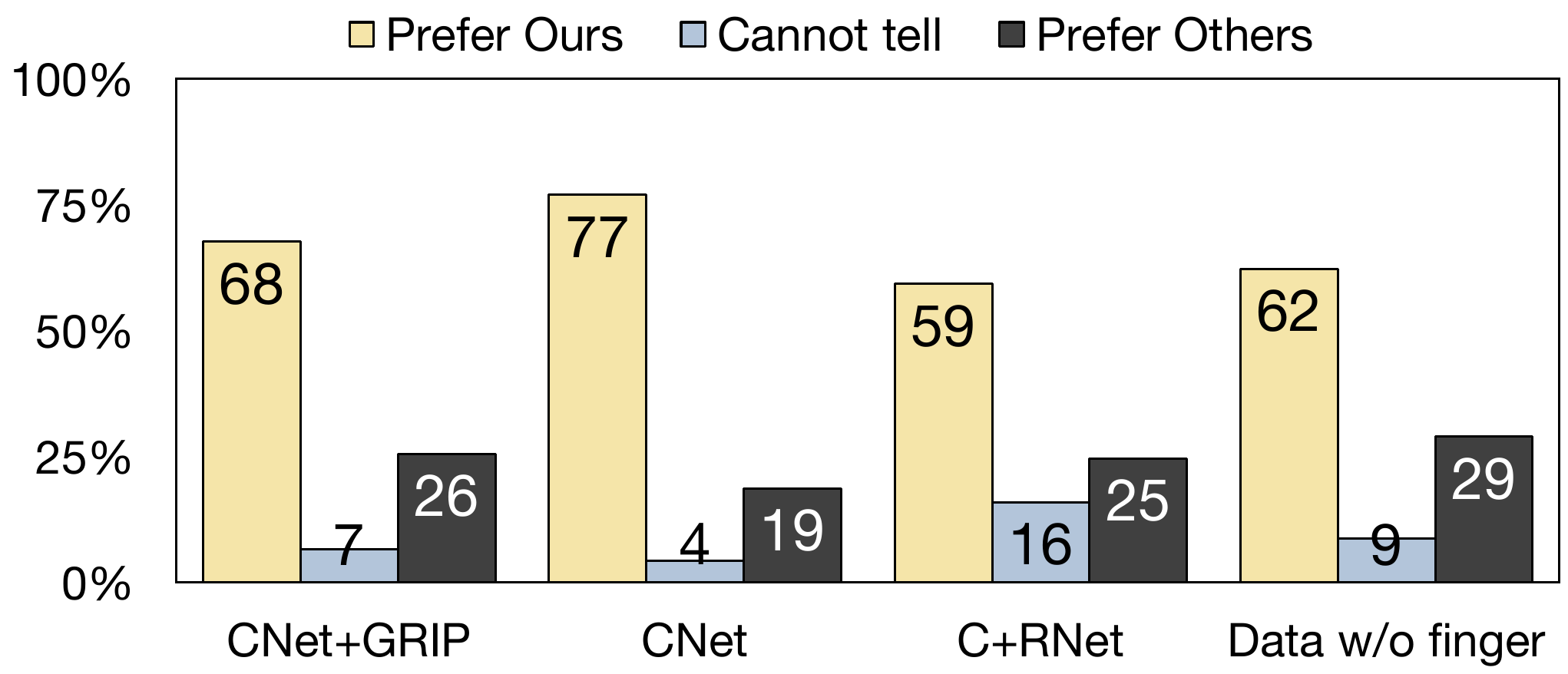}
\vspace{-5mm}
    \caption{Comparison of our full model and baseline models through human perception studies.}
    \label{fig:interaction_human_study}
\vspace{-6mm}
\end{figure}

\setlength{\columnsep}{10pt}
\begin{wraptable}{r}{0.45\columnwidth}
\centering
\vspace{-10pt}
\caption{Comparison of geometric accuracy between our method and the baseline.}
\begin{adjustbox}{max width=0.45\columnwidth}
\begin{tabular}{lcc}
\toprule
& PE$_\text{p} \downarrow$ & PE$_\text{o} \downarrow$ \\
\midrule
Baseline & 21.9\% & 12.5\% \\
Ours & \textbf{3.1\%} & \textbf{1.6\%} \\
\bottomrule
\end{tabular}
\end{adjustbox}
\label{tab:llm_geometry}
\vspace{-10pt}

\end{wraptable}
For geometric accuracy, we measure the proportion of objects with positional (PE$_\text{p}$) or orientation errors (PE$_\text{o}$) in the scene. 
Positional errors include objects placed at incorrect heights or with penetration into other objects, while orientation errors refer to objects being directed in the wrong direction.
The results, shown in \cref{tab:llm_geometry}, demonstrate that our method outperforms the baseline by a large margin. A qualitative comparison is presented in the upper part of \cref{fig:llm_result}, where the baseline generates incorrect object positions and orientations, while our method produces more reasonable layouts.

We also conducted a human perception study, with results presented in the lower part of \cref{fig:llm_result}. Each pair was reviewed by 20 Amazon Mechanical Turk (AMT) workers. They were asked to choose the scene that was more reasonable and consistent with the given instruction. The majority of participants preferred our method.

\subsection{Evaluation of Low-Level Motion Generator}
\label{subsec:low_level_eval}

\paragraph{Datasets.}
CoarseNet and RefineNet are trained on the FullBodyManipulation dataset~\cite{li2023object}, which includes 10 hours of human-object interaction with 15 objects but lacks finger motion.
FingerNet uses the GRAB dataset~\cite{GRAB:2020}, containing full-body and finger motion for 10 subjects with 51 objects.
The navigation module is trained on HumanML3D~\cite{guo2022generating}, featuring 28 hours of diverse motions with language descriptions.
For all datasets, we follow standard data partitioning from prior work~\cite{li2023object, GRAB:2020, guo2022generating}.

\vspace{-4mm}
\paragraph{Evaluation Metrics.}
Following prior work~\cite{li2023controllable}, we evaluate our approach from various aspects.
All distance errors are measured in centimeters. 
For condition matching, we report the waypoint following errors ($T_{xy}$).
For human motion quality, we report the foot sliding score (FS) and foot height ($H_{\text{feet}}$).
For interaction quality, we use contact precision (C\textsubscript{prec}), recall (C\textsubscript{rec}), and F1 score (C\textsubscript{F1}).
To better assess finger-object interaction, we also report contact coverage (CC) -- the percentage of hand points within -2mm to +2mm of the object surface -- and intersection volume (IV), the overlap volume of hand and object meshes in $\text{cm}^3$, following~\cite{grady2021contactopt}.
For ground truth difference, we report the mean per-joint position error (MPJPE), object position error ($T_{\text{obj}}$), and the object orientation error ($O_{\text{obj}}$).
We also conduct a human perception study to measure the motion quality.

\vspace{-4mm}
\paragraph{Baselines.}
For generating full-body human motion, finger motion, and object motion from text, prior work~\cite{ghosh2022imos,li2024task} trained models on GRAB~\cite{GRAB:2020} to predict motions for small-object manipulation. However, GRAB contains only small-object manipulation data without locomotion, so models trained on it cannot generalize to scenarios involving large objects that require full-body coordination and locomotion. Moreover, the approach from~\cite{ghosh2022imos,li2024task} is not applicable to the FullBodyManipulation dataset~\cite{li2023object} used in our work, as this dataset lacks finger motions, making it infeasible to learn a model via direct supervision. 
Due to these differences, a directly comparable baseline does not exist.
Thus, we establish our baseline by combining two complementary methods: CHOIS~\cite{li2023controllable} (referred to as CoarseNet in our framework), which generates object motion and full-body motion from text, and GRIP~\cite{taheri2023grip}, which refines arm movements and generates finger motions based on CHOIS's output.

In addition, to evaluate the impact of each component, we compare our full system against two ablations: (1) CoarseNet alone (CNet) and (2) CoarseNet + RefineNet (C+RNet), which represents the full system without finger motion. 
The full system is denoted as C+R+FNet.
This evaluation focuses on the motion synthesis quality before physics tracking. We present the evaluation for the physics tracker in \cref{subsec:phys_eval}.

\vspace{-4mm}
\paragraph{Results.}
The qualitative comparison in \cref{fig:interaction_qualitative} clearly demonstrates the superiority of our method. Both ablated variants show a lack of finger detail, and GRIP~\cite{taheri2023grip} fails to generate accurate hand motion, resulting in significant artifacts. In contrast, our method produces precise finger-object interactions.
We also observe the impact of RefineNet: compared to CNet, C+RNet maintains more consistent and reasonable relative poses between the hand and object during the interaction, showing RefineNet’s role in improving pose consistency and reducing artifacts.

The quantitative results are presented in \cref{tab:interaction}.
Since our method focuses on finger-object interaction, the five metrics under the \textit{Interaction} column are particularly important, and we achieve the best performance across all of them.

We further conduct a human perception study to complement our evaluation. We generate 15 sequences using each method and compose 60 pairs, with each pair consisting of our results and the results of a baseline. Each pair was reviewed by 20 Amazon Mechanical Turk (AMT) workers. As shown in \cref{fig:interaction_human_study}, our method is consistently preferred over the baselines by a large margin. Notably, our method achieves even higher preference rates when compared to the real-world data from the FullBodyManipulation~\cite{li2023object} dataset, as we generate detailed finger motions that are absent in the original dataset.

\subsection{Evaluation of Physics Tracker}
\label{subsec:phys_eval}

\begin{figure}[t]
    \centering
\includegraphics[width=0.48\textwidth]{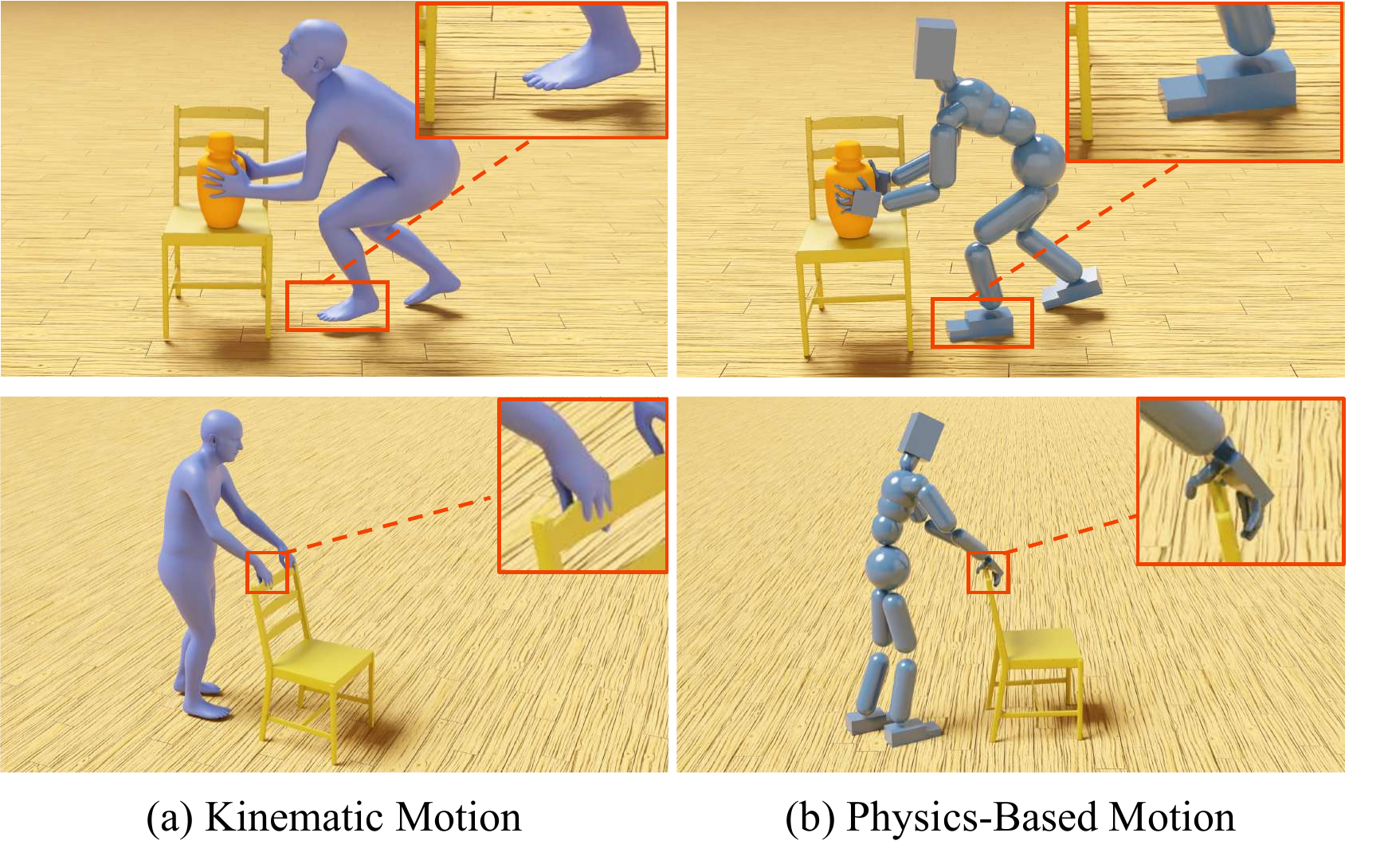}
\vspace{-7mm}
    \caption{Qualitative comparison of kinematic motion and physics-based motion. The physics tracker corrects artifacts such as foot floating and penetration.} 
    \label{fig:physics_qualitative_results}
\end{figure}

The physics tracker can accurately track kinematic motion while ensuring the physical plausibility of the motion. Through physics simulation, artifacts such as foot floating, foot sliding, and human-object penetration are eliminated, as shown in \cref{fig:physics_qualitative_results}. 

\setlength{\columnsep}{10pt}
\begin{wraptable}{r}{0.4\columnwidth}
\centering
\vspace{-1pt}
\caption{Tracking error between the kinematic motion and physics-based motion.}
\begin{adjustbox}{max width=0.35\columnwidth}
\begin{tabular}{lcc}
\toprule
& $E_h$ & $E_o$ \\
\midrule
Ours & 5.45 & 4.67 \\
\bottomrule
\end{tabular}
\end{adjustbox}

\label{tab:physics}
\vspace{-6pt}
\end{wraptable}
Additionally, we evaluate the tracking error of human joints ($E_h$) and objects ($E_o$), as shown in \cref{tab:physics}. The tracking error is defined as the positional error (cm) between the kinematic motion and the tracked motion. 
The results were evaluated across 20 interaction sequences, each with an average duration of 30 seconds and 2 objects of varying shapes and types.
All sequences were successfully tracked, with relatively small errors, demonstrating high tracking accuracy.

%% file: chapters/9_conclusion.tex
\section{Conclusions}
In this work, we introduced a system that simulates realistic human-object manipulation in a contextual environment based on human-level language instructions.
The framework consists of a high-level LLM planner that infers a target scene layout and execution plan from the given instructions, a low-level motion generator that generates synchronized, full-body, fingers, and object motion, and a physics tracker that imitates the generated kinematic motions and produces physically plausible interaction motions.
Our system effectively handles a variety of tasks, showcasing its potential for various real-world applications.

\paragraph{Future Work.}
Our system currently uses text or image inputs to represent the contextual environment, and could potentially be enhanced with more structured representations like voxel grids as used in \cite{jiang2024autonomous}. 
Building on this idea, a promising direction is to create an agent with egocentric vision perception as input to LLM/VLMs for high-level motion planning. 
This could enable the agent to exhibit more natural behaviors, such as purposeful head orientation for observation and better body positioning to avoid collisions.



%% file: chapters/10_ack.tex
\paragraph{Acknowledgement.}
We thank Wenyang Zhou for data processing. This work is in part supported by the Wu Tsai Human Performance Alliance at Stanford University and the Stanford Institute for
Human-Centered AI (HAI). 

%% file: supp.tex
\renewcommand{\thefigure}{S\arabic{figure}}
\renewcommand{\thetable}{S\arabic{table}}
\def\theequation{S\arabic{equation}}
\setcounter{page}{1}
\setcounter{section}{0}
\setcounter{equation}{0}
\setcounter{table}{0}
\setcounter{figure}{0}
\maketitlesupplementary

\renewcommand{\thesection}{\Alph{section}}
\renewcommand{\thesubsection}{\thesection.\arabic{subsection}}
\setcounter{tocdepth}{2}

\section{Details of High-Level Planner}

\subsection{Pseudo-code for Object Positions Calculation}

In \cref{alg:compute_position}, we provide the pseudo-code for the algorithm that calculates object positions in the scene graph, as described in~\cref{subsec:spatial_relationship}.

\begin{algorithm}
\SetAlgoLined
\DontPrintSemicolon
\SetKwInOut{Input}{Input}
\SetKwInOut{Output}{Output}

\Input{Scene graph $G(V, E)$ with vertices $V$ and edges $E$, initial positions $L$, static nodes ($V_{S}$), the nodes to be moved ($V_{M}$).}
\Output{Updated positions $L$ for all nodes.}

\SetKwFunction{FSub}{ComputePositions}
\SetKwFunction{FUpdate}{Update}
\SetKwFunction{FOffset}{ComputeOffset}

\SetKwProg{Fn}{Procedure}{:}{\KwRet}
\Fn{\FSub{$G$, $L$}}{
    Let the set of processed nodes $V_{P} \gets V_{S}$\;
    
    \While{$V_{M}$ is not empty}{
        Let \( V' \subseteq V_M \) be the nodes whose predecessors' positions are known\;
        \ForEach{$v \in V'$}{
            $L(v) \gets$ \FUpdate{$v$, $G$, $L$}\;
            $V_{P} \gets V_{P} \cup \{v\}$\;
            $V_M \gets V_M \setminus \{v\}$
        }
    }
    \KwRet $L$\;
}

\Fn{\FUpdate{$v$, $G$, $L$}}{
    Initialize an empty list of positions $P$\;
    \ForEach{predecessor $u$ of $v$ in $G$}{
        $O \gets$ \FOffset{$(u, v)$, $L(u)$}\;
        Append $L(u) + O$ to $P$\;
    }
    \KwRet $\text{Average}(P)$\;
}

\Fn{\FOffset{$(u, v)$, $L(u)$}}{
    \If{edge $(u, v)$ indicates ``on''}{
        Set $O_{\text{height}}$ to align with the top surface of $u$\;
        Sample $O_{\text{horizontal}}$ within the horizontal bounds of $u$\;
        \KwRet $[O_{\text{horizontal}}, O_{\text{height}}]$\;
    }
    \ElseIf{edge $(u, v)$ indicates ``adjacent''}{
        Use the direction and distance provided by the LLMs to compute $O$\;
        \KwRet $O$\;
    }
}

\caption{Object Positions Calculation}
\label{alg:compute_position}
\end{algorithm}

\subsection{Prompts for High-Level Planner}
\begin{figure*}[h]
    \centering
\includegraphics[width=\textwidth]{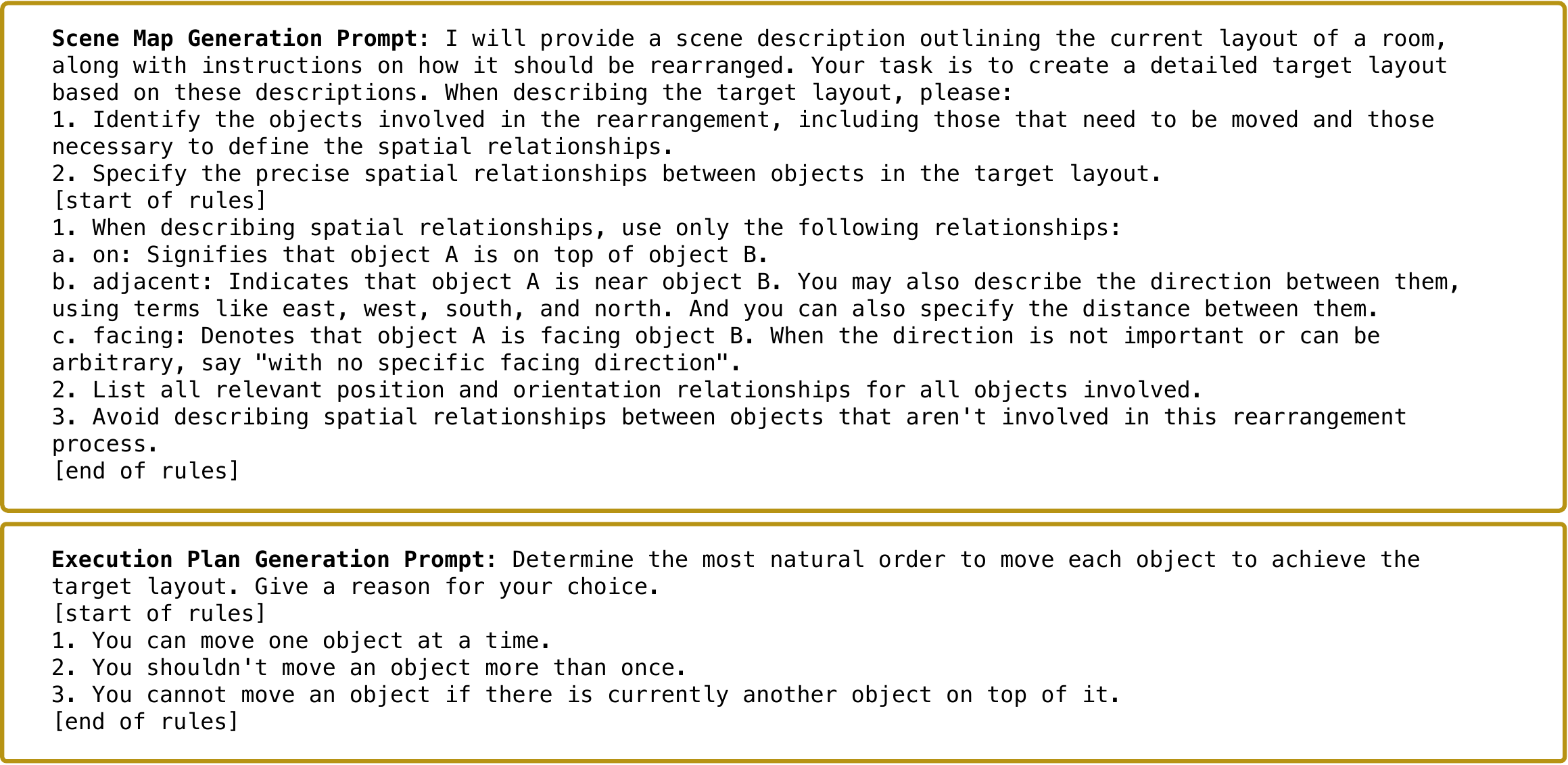}
    \caption{Prompts used by the LLM planner to generate the scene map and execution plan.} 
    \label{fig:llm_spatial_text}
    \vspace{-2mm}
\end{figure*}

\begin{figure*}[h]
    \centering
\includegraphics[width=\textwidth]{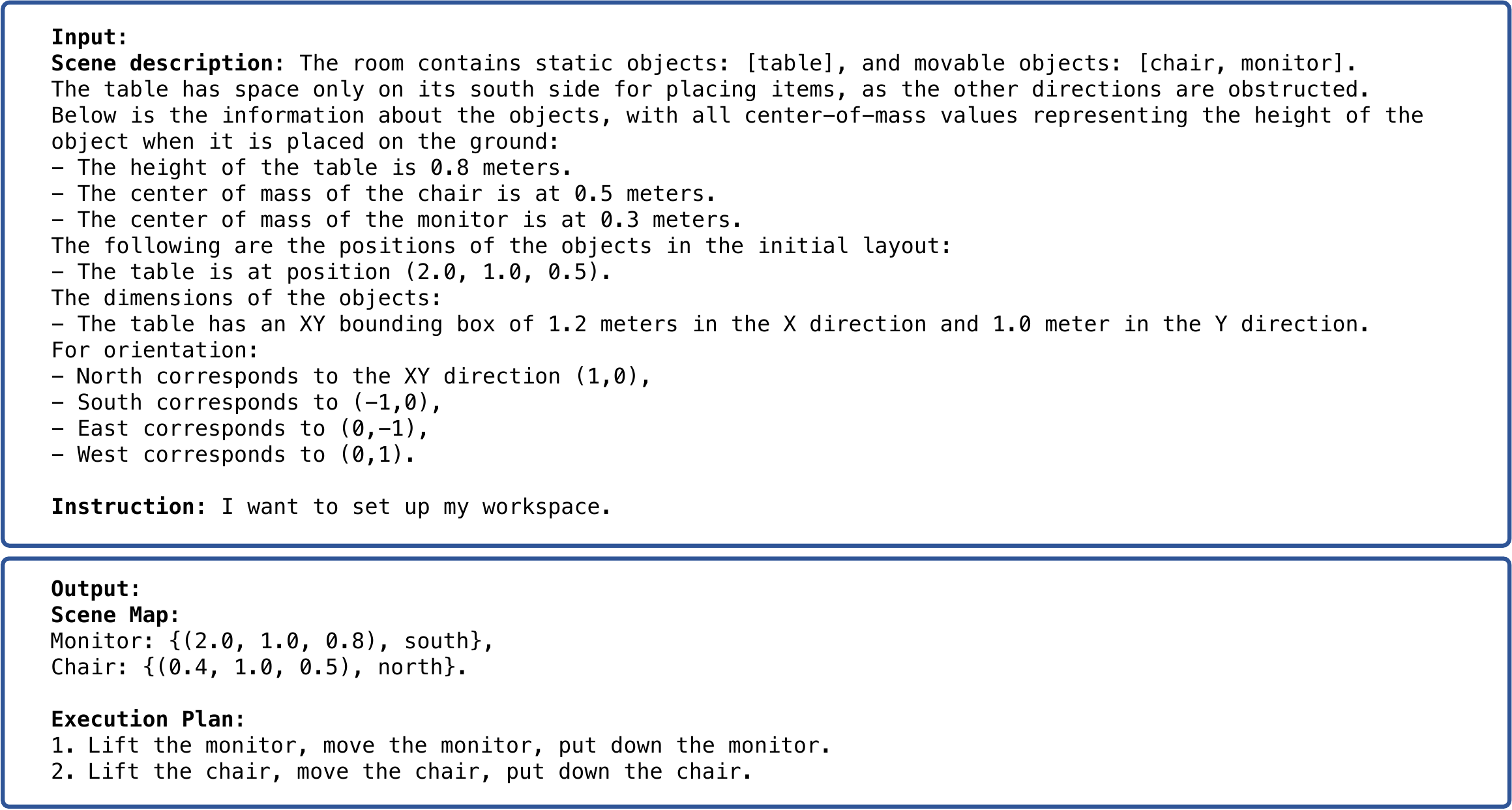}
    \caption{Example of the input and output for the LLM planner.} 
    \label{fig:llm_our_example}
    \vspace{-2mm}
\end{figure*}

\begin{figure*}[h]
    \centering
\includegraphics[width=\textwidth]{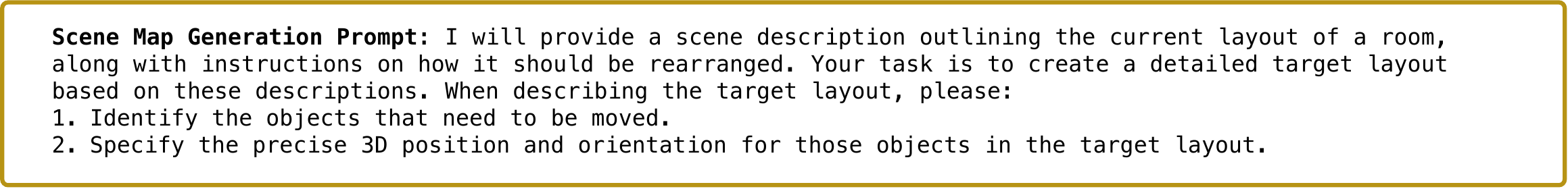}
    \caption{Prompts used by the baseline, as described in~\cref{subsec:planner_eval}.} 
    \vspace{-2mm}
    \label{fig:llm_baseline}
\end{figure*}

We provide the prompts used for generating the scene map and execution plan in~\cref{fig:llm_spatial_text}. 
An example of the input and the corresponding output of the LLM planner is shown in~\cref{fig:llm_our_example}.
We also present the prompt used by the baseline (as described in~\cref{subsec:planner_eval})
in~\cref{fig:llm_baseline}.

\section{Details of Low-Level Motion Generator}
\subsection{Diffusion Model Architecture}
\label{supp:diff_arch}

\begin{figure}[!t]
    \centering
\includegraphics[width=0.48\textwidth]{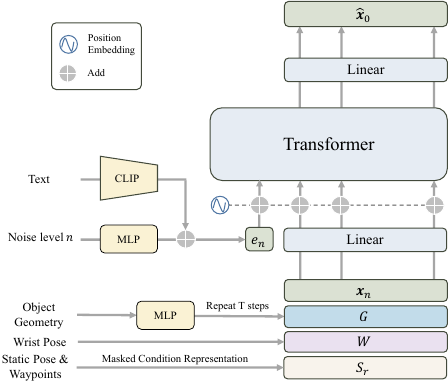}
    \caption{Architecture of RefineNet. The model is conditioned on $\bm{c}_{r} = \left\{\bm{W}, \bm{S}_{r}, \bm{G}, \bm{T}\right\}$, where $\bm{W}$ represents the wrist-object relative pose, $\bm{S}_{r}$ is the masked motion representation containing the static pose and waypoints, $\bm{G}$ denotes the object geometry, and $\bm{T}$ is the textual input.}
    \label{fig:supp_diffusion}
    
\end{figure}

Our system includes four diffusion models: CoarseNet, RefineNet, FingerNet, and the navigation module, all of which adopt a similar transformer-based architecture. Here, we illustrate the architecture of RefineNet in \cref{fig:supp_diffusion}. The model is provided with a motion sequence $\bm{x}_n$, the condition $\bm{c}_r$, and the noise step $n$, and predicts the clean motion $\hat{\bm{x}}_{0}$.
For the other networks, we simply substitute the condition with the corresponding one.

\subsection{Implementation Details of Grasp Pose Generation}
As introduced in~\cref{subsubsec:grasp_pose}, we use a state-of-the-art grasp pose generation
method ~\cite{wang2023dexgraspnet} to generate the grasp pose.
Although originally designed for robotic hand grasping, this method relies solely on kinematic chain information, making it fully compatible with the SMPL-X~\cite{smplx} hand model. Starting with an object mesh and an initial hand pose, the grasp generation process is formulated as an optimization problem that minimizes a differentiable force closure energy term along with several additional terms that encourage the fingers to conform closely to the object surface while maintaining a natural hand shape. For motions involving both hands, such as carrying a box, we omit the force closure term.

\subsection{Post-processing in RefineNet}
\label{subsec:post_process}
As introduced in~\cref{sec:refine_net}, we apply post-processing to RefineNet's output to better align it with the input condition.
First, we replace the object poses in the pre-contact and post-contact phases with the corresponding static poses. To address the abrupt transitions at the phase boundaries, we identify the difference between the predicted object pose and the static object pose at the boundary frame denoted as $\Delta \bm{d} \in \mathbb{R}^6$ (suppose this is the boundary between pre-contact phase and contact phase), which encapsulates both the positional and rotational differences in axis-angle format. To smooth these transitions, we employ an interpolation strategy to the object motion in the contact phase. The interpolation function is defined by $\hat{\bm{O}}_{t} = \bm{O}_{t} + \alpha_t \Delta \bm{d}$, where $\alpha_t$ is a time-dependent scaling factor that decreases from 1 to 0 across the transition period, $\bm{O}_t$ denotes the predicted object pose. 

Next, we use the rectified object motion and the optimized wrist pose $\hat{\bm{w}}=(\bm{\hat{R}}, \bm{\hat{T}})$ to compute the wrist trajectories during the contact phase. 
The rotation is computed as $\bm{R}_o\bm{\hat{R}}$, and the position is computed as $\bm{R}_o \bm{\hat{T}} + \bm{T}_o$, where $\bm{R}_o$ and $\bm{T}_o$ represent the object's rotation and position, respectively.
The computed wrist trajectories may slightly deviate from the predicted wrist trajectories in the contact phase, leading to unsmooth transitions near the phase boundaries. We apply the same interpolation strategy to the wrist poses in the pre-contact and post-contact phases to produce smooth wrist motions. Finally, we employ an Inverse Kinematics (IK) algorithm to solve for the updated human pose constrained by the wrist poses.

\subsection{Implementation Details of FingerNet}
FingerNet generates finger motions for the pre-contact and post-contact phases to create natural transitions between rest and grasp poses. It is trained on the GRAB dataset~\cite{GRAB:2020}. Since FingerNet focuses solely on the hand approaching and releasing motions, we only use the corresponding portions of the dataset. Specifically, we extract one second of motion before the grasp begins and after it ends as training data, discarding the rest.

We adopt a mirroring strategy similar to~\cite{zhang2021manipnet} to handle both hands simultaneously. During training, left-hand data are mirrored and combined with the right-hand data so that the model only needs to predict motions for the right hand. At inference time, we mirror the left-hand input, predict the corresponding motions using the model, and then mirror the output back to obtain the left-hand motions.

\subsection{Smooth Transition between Navigation and Interaction}
The navigation and interaction modules alternate to generate a long-horizon interaction sequence. Specifically, when the human is far from the object, the navigation module is activated to guide the human towards it. Once the human is within a threshold distance (set at 1 meter), the system switches to the interaction module to begin object manipulation. After the interaction is completed and the object is put down, control reverts to the navigation module to move towards the next object.

Since both modules use the initial human pose as an input condition, we ensure a smooth transition by feeding the last pose from the previous module into the next.
We apply the same smoothing technique described in \cref{subsec:post_process} at the switching boundary to further enhance continuity.

\section{Implementation Details of Physics Tracker}
We use IsaacGym~\cite{makoviychuk2021isaac} as the physics engine to perform simulation.
The simulated character is skeleton-driven where each joint is modeled as a motor controlled through a PD controller.
The control policy works at 30Hz, which is the same with generated kinematics motions, while the simulation runs at 120Hz.
Our character has $62$ body links and $49$ controllable joints.
This results in a state space $\mathbf{s}_t \in \mathbb{R}^{(62+|\mathcal{O}|)\times 13}$ including the position, orientation and linear and angular velocities of the character's body links and the manipulating objects $\mathcal{O}$ depending on the given tracking motions.
To track the target motion,
additional observation $\mathbf{o}_t \in \mathbb{R}^{(62+|\mathcal{O}|)\times7}$ including the target position and orientation of each body link and object is also introduced as the input to the control policy.
The action space is $\mathbf{a}_t \in \mathbb{R}^{49\times3}$ given each controllable joint has 3 degrees of freedom.
All our control policies are optimized using PPO~\cite{schulman2017proximal} as the backbone reinforcement learning algorithm with network parameters updated by Adam optimizers~\cite{kingma2014adam}.
The hyperparameters used for policy training are listed in Tab.~\ref{tab:rl_hyper}.

\subsection{Reward Design}
Since fingers take $30$ out of $49$ controllable joints and have a higher requirement in control precision for accurate object manipulation, we define a reward function containing three terms to evaluate the full-body pose and finger pose separately:
\begin{equation}
    r = 0.8 r_\text{body} + 0.2 r_\text{hand} + 0.05 r_\text{energy},
\end{equation}
where $r_\text{body}$ evaluates the full-body poses excluding fingers, $r_\text{hand}$ evaluates the hand poses related to the target manipulating object,
and $r_\text{energy}$ is an additional energy-related term penalizing end effectors' abrupt movement.
Note that in our implementation, at any time step,
at most one object is activated as the manipulation target, while multiple objects may exist in the scene and would be manipulated one by one.

\begin{table}[t]
    \centering
    \begin{tabular}{rl}
    \toprule
       \textbf{Body Part}  &  \quad\textbf{Value} \\
       \midrule
       root (pelvis) & $w_q = 1, w_p = 1$\\
       lower abdomen & $w_q = 0.2$\\
       upper abdomen & $w_q = 0.2$\\
       chest & $w_q = 0.2$\\
       neck & $w_q = 0.2$\\
       head & $w_q = 0.2$\\
       clavicles & $w_q = 0.1$\\
       upper arms & $w_q = 0.2$ \\
       lower arms & $w_q = 0.2$\\
       wrists & $w_q = 0.3, w_p = 0.3$\\
       thighs & $w_q = 0.5$\\
       calfs & $w_q = 0.3$\\
       feet & $w_q = 0.2, w_p = 0.1$\\
       target object & $w_q = 1, w_p = 1$ \\
        \bottomrule
    \end{tabular}
    \caption{Unnormalized weights of each body link of the character and the target object when computing the reward regarding body pose errors (cf. $r_\text{body}$ in Equation~\ref{eq:r_body}). All weights unlisted are zero.
    The weights are normalized such that $\sum_b w_{p,b}=1$ and $\sum_b w_{q,b}=1$ before computing $r_\text{body}$.}
    \label{tab:rl_weights}
\end{table}

\begin{table}[t]
    \centering
\begin{tabular}{lc}
    \toprule
    \textbf{Parameter} & \textbf{Value}\\
    \midrule
    policy network learning rate & $5 \times 10^{-6}$\\
    critic network learning rate & $1 \times 10^{-4}$\\
    reward discount factor ($\gamma$) & $0.95$ \\
    GAE discount factor ($\lambda$) & $0.95$ \\
    surrogate clip range ($\epsilon$) & $0.2$ \\
    number of PPO workers & \\
    \quad (simulation environment instances) & $2048$ \\
    PPO replay buffer size & $2048\times8$ \\
    PPO batch size & $256$ \\
    PPO optimization epochs & $5$ \\
  \bottomrule
  \end{tabular}
    \caption{Hyperparameters of policy training for physics tracker.}
    \label{tab:rl_hyper}
    
\end{table}

For full-body pose evaluation,
we consider the orientation and position accuracy of the character's body links and the target object:
\begin{equation}\label{eq:r_body}\begin{split}
    r_\text{body} = & 0.5\exp\left(-15 \sum_{b\in\mathcal{B}} w_{q,b} ||\mathbf{e}_{q,b}||^2\right) \\
    & \quad + 0.5\exp\left(-15 \sum_{b\in\mathcal{B}} w_{p,b} ||\mathbf{e}_{p,b}||^2\right),
\end{split}
\end{equation}
where $\mathcal{B}$ is the set of all body links of the character and the activated target object at the current time step, $\mathbf{e}_{q,b}$ is the orientation error between each body link $b$ of the simulated character and that in the tracking motion, $\mathbf{e}_{p,b}$ is the position error, and $w_{q,b}$ and $w_{p,b}$ are the associated weights.
We measure $\mathbf{e}_{q,b}$ using the radian of the angle needed to rotate the link $b$ to the target orientation,
while $\mathbf{e}_{p,b}$ is measured by the Euclidean distance from the current position of link $b$ to the target position.
All weight values are listed in Tab.~\ref{tab:rl_weights}. For position tracking, we focus only on the position of the root link and end effectors to ensure that the trajectory of end effectors (foot and hand excluding fingers) matches the tracking target,
while relying on orientation tracking to ensure the naturalness of full-body poses.

For hand pose evaluation,
to ensure that the object would be manipulated by hands correctly,
we consider the fingers' relative position to the target manipulating object when the hand is close to the object, or their relative position to the wrist when the hand is far away from the target object:
\begin{equation}
    r_\text{hand} = \exp\left(- \frac{5}{|\mathcal{F}|}\sum_{f\in\mathcal{F}} \alpha ||\mathbf{e}_{f,o}|| + (1-\alpha)  ||\mathbf{e}_{f,w}||\right),
\end{equation}
where $\mathcal{F}$ is the set of fingers, $\mathbf{e}_{f,o}$ is the position error between each finger $f$ of the simulated character and the tracking target position in the object's local system, $\mathbf{e}_{f,w}$ is the position error in the wrist's local system, and $\alpha$ is an interpolation coefficient depending on the distance between the hand and object in the generated tracking motions. We define $\alpha=1$ when the hand is equal to or less than 0.25m from the target object and $\alpha=0$ if the hand is at least 1m far away from the object in the tracking motion.

$r_\text{energy}$ takes into account the character's end effectors to prevent jittering or abrupt movement with larger acceleration:
\begin{equation}
    r_\text{energy} = \exp\left(-\frac{1}{900}\sum_{e\in\mathcal{E}} ||\mathbf{a}_e||^2\right),
\end{equation}
where $\mathcal{E}$ is the set of the character's key end effectors (feet and hands excluding fingers), and $\mathbf{a}_e$ is the linear acceleration of the end effector $e$.

In contrast to recent works~\cite{brahmbhatt2019contactgrasp,wang2023physhoi,luo2024grasping} on object manipulation, our reward function does not rely on the contact states between the hands (or fingers) and objects. This is because the contact data derived from the kinematic motions generated by diffusion models can be unreliable due to artifacts (cf.~\cref{fig:physics_qualitative_results}). Moreover, this approach aligns with the common scenario in which detailed contact information is typically absent in motions obtained from motion capture and video demonstrations~\cite{pollard2005physically}.

\subsection{Importance Sampling Strategy}

Given that the generated kinematic motions can span several minutes and involve interactions with multiple objects, it would be highly inefficient to perform motion tracking from the very beginning of the sequence. Since locomotion without object manipulation and the phase following object grasping are relatively easier to learn, we adopt an importance sampling strategy that focuses on pre-grasp states to facilitate policy training.
Specifically, we partition the parallel simulated physics environments into $|\mathcal{O}|$ batches, with each batch assigned to a specific manipulating object. In each batch, simulated characters are initialized in a random pre-grasp pose while interacting with the assigned object. This approach increases the likelihood that the policy learns the transition from the pre-grasp to the grasping phase, and subsequently, the locomotion that follows a successful grasp.
Moreover, this strategy ensures that the policy consistently learns to interact with multiple objects simultaneously, thereby avoiding potential local minima during policy exploration.

\section{Inference Speed}
Achieving real-time performance remains an open challenge for both diffusion-based and optimization-based methods. In our low-level motion generator, producing a 4-second motion sequence takes about four seconds for each diffusion model in stages 1, 3, and 4, while the optimization process in stage 2 requires roughly three minutes. Additionally, training the physics tracker on a 4-second interaction segment takes approximately 2 hours.
All experiments were run on a single NVIDIA RTX 4000 GPU.

\begin{figure}[t]
    \centering
\includegraphics[width=0.48\textwidth]{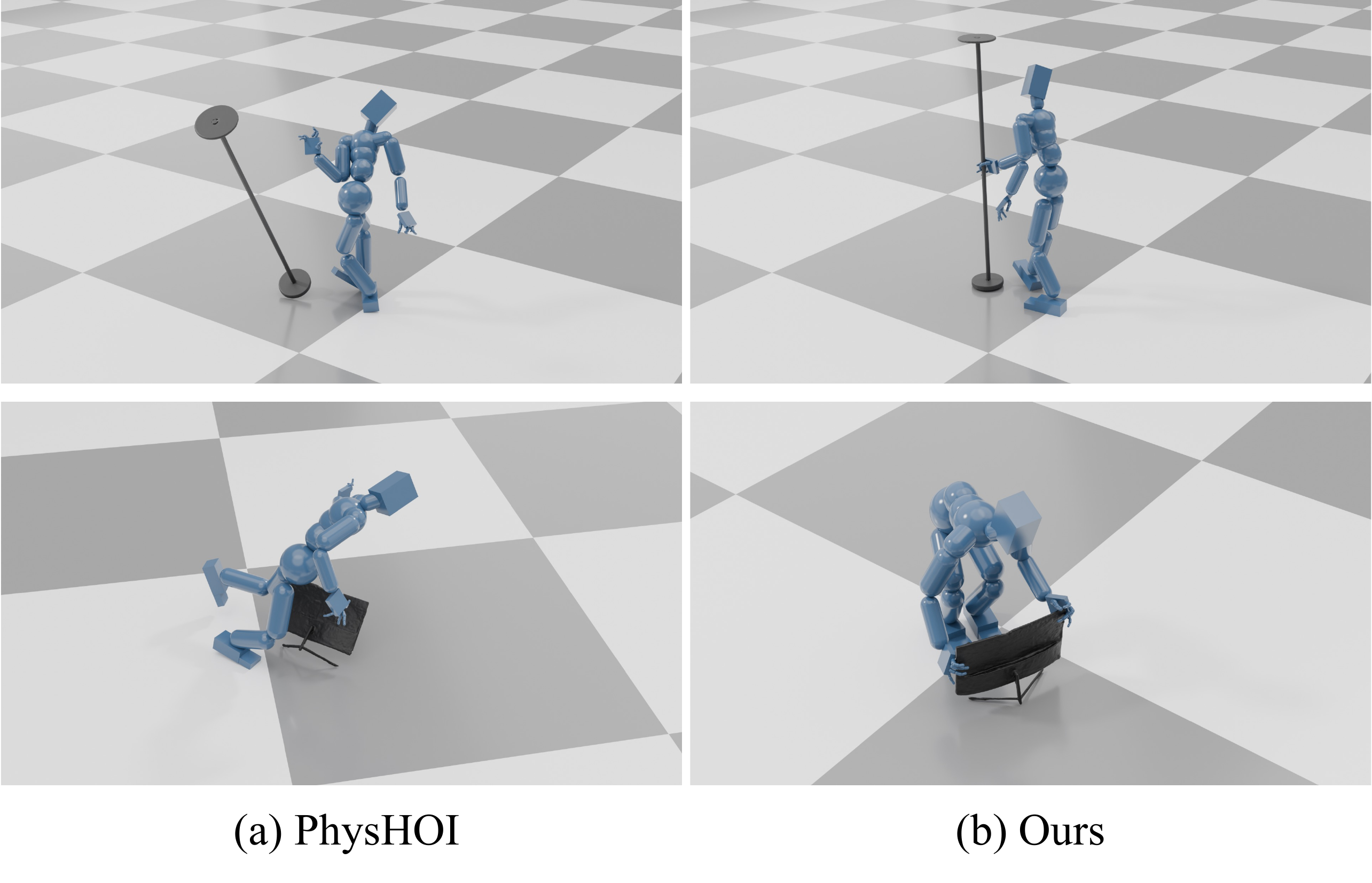}
    \caption{PhysHOI~\cite{wang2023physhoi} fails to track kinematic motion, while our policy successfully drives the character to interact with diverse objects.}
    \label{fig:physhoi}
\end{figure}

\section{Additional Experiemental Analysis}

\subsection{Implementation Details of Baseline}
Since no prior work addresses the task of generating full-body motion, finger motion, and object motion from language instructions, we adapt prior works CHOIS~\cite{li2023controllable} and GRIP~\cite{taheri2023grip} as fair baselines to compare against our system. CHOIS generates full-body motion and object motion from text and initial states, which also serves as our Stage 1. The output of CHOIS will serve as input to GRIP. GRIP consists of an arm denoising model and a two-stage finger motion generation model. It takes full-body motion (without finger motion) and object motion as input to denoise arm motion. It then proposes a hand sensor to extract hand-object spatial features and predict initial finger poses. In the second stage of finger motion generation, GRIP leverages these predicted finger poses to compute proximity features, which serve as input to refine the finger motion.

\subsection{Results of Navigation Module}
We report the evaluation scores for the navigation module in \cref{tab:navigation}, using the metrics introduced in~\cref{subsec:low_level_eval}. The module is evaluated on the test split of the HumanML3D dataset \cite{guo2022generating}. It demonstrates that the module closely adheres to the input waypoints, with a relatively small joint position error when compared to the ground truth data, while maintaining reasonable foot sliding and foot height.

\begin{table}[t]
\centering
\begin{adjustbox}{max width=0.5\textwidth}
\begin{tabular}{lcccccccccccccccccc}
\toprule
& $T_{xy}$ & $H_{\text{feet}}$ & $FS$ & $\text{MPJPE}$  \\
\midrule

Ours
& 4.00 & 1.48 & 0.58 & 10.87 \\
\bottomrule
\end{tabular}
\end{adjustbox}

\caption{Navigation synthesis on the HumanML3D dataset \cite{guo2022generating}. The joint error is relatively small, demonstrating that the model aligns well with the input waypoints.}
\label{tab:navigation}
\end{table}

\subsection{Results of Physics Tracker}
We compare our physics tracker against PhysHOI~\cite{wang2023physhoi}, which trains reinforcement learning policies to control a humanoid character playing basketball. 
For our task, we have PhysHOI track the kinematic motion generated by our low-level motion generator. As shown in \cref{fig:physhoi}, PhysHOI fails to grasp objects, whereas our method successfully completes the task.

\subsection{Human Perception Study Details}
\begin{figure*}[h]
    \centering
\includegraphics[width=\textwidth]{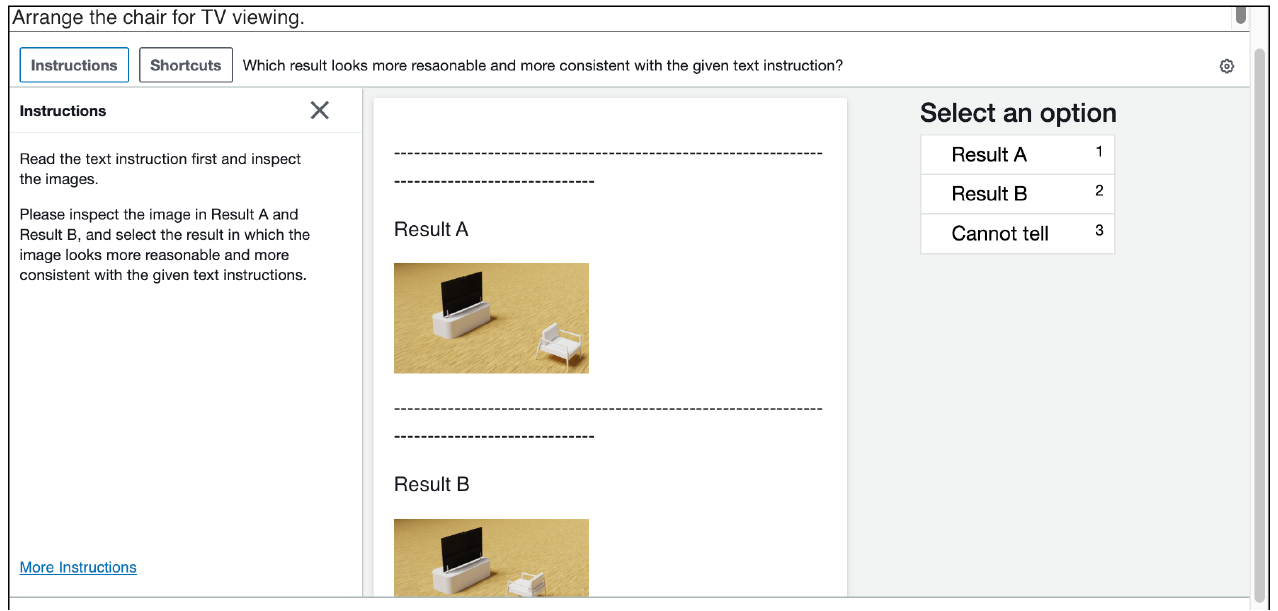}
    \caption{Interface of human perceptual study for the high-level planner.} 
    \label{fig:llm_human_study}
    \vspace{-2mm}
\end{figure*}

\begin{figure*}[h]
    \centering
\includegraphics[width=\textwidth]{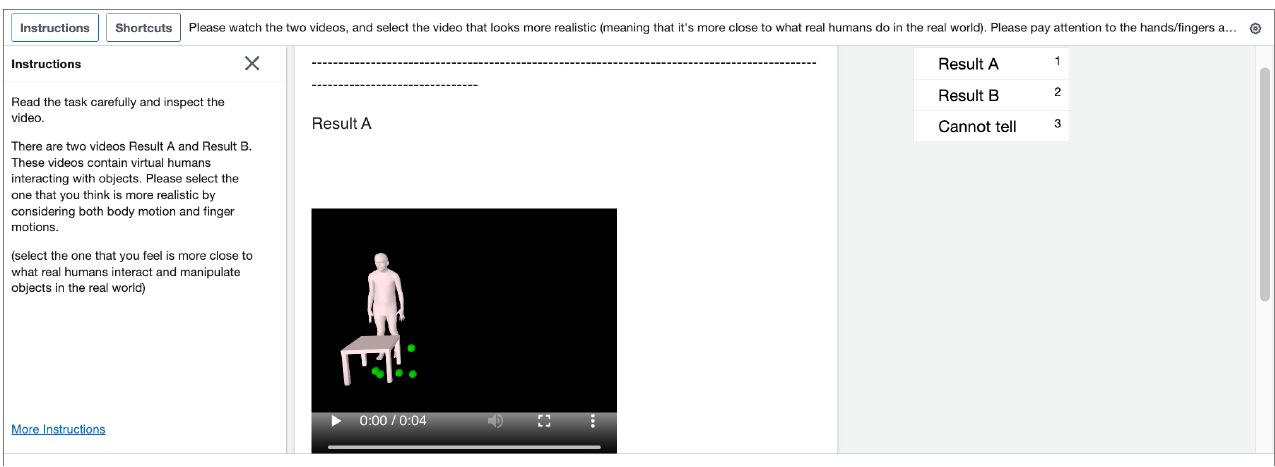}
    \caption{Interface of human perceptual study for the low-level motion generator.} 
    \label{fig:motion_human_study}
    \vspace{-2mm}
\end{figure*}

We show the interface of the human perceptual study for the high-level planner in Figure~\ref{fig:llm_human_study} and the low-level motion generator in Figure~\ref{fig:motion_human_study}.






%% file: main.bbl
\begin{thebibliography}{74}
\providecommand{\natexlab}[1]{#1}
\providecommand{\url}[1]{\texttt{#1}}
\expandafter\ifx\csname urlstyle\endcsname\relax
  \providecommand{\doi}[1]{doi: #1}\else
  \providecommand{\doi}{doi: \begingroup \urlstyle{rm}\Url}\fi

\bibitem[Aguina-Kang et~al.(2024)Aguina-Kang, Gumin, Han, Morris, Yoo, Ganeshan, Jones, Wei, Fu, and Ritchie]{aguina2024open}
Rio Aguina-Kang, Maxim Gumin, Do~Heon Han, Stewart Morris, Seung~Jean Yoo, Aditya Ganeshan, R~Kenny Jones, Qiuhong~Anna Wei, Kailiang Fu, and Daniel Ritchie.
\newblock Open-universe indoor scene generation using llm program synthesis and uncurated object databases.
\newblock \emph{arXiv preprint arXiv:2403.09675}, 2024.

\bibitem[Araujo et~al.(2023)Araujo, Li, Vetrivel, Agarwal, Gopinath, Wu, Clegg, and Liu]{araujo2023circle}
Joao~Pedro Araujo, Jiaman Li, Karthik Vetrivel, Rishi Agarwal, Deepak Gopinath, Jiajun Wu, Alexander Clegg, and C~Karen Liu.
\newblock Circle: Capture in rich contextual environments.
\newblock In \emph{Conference on Computer Vision and Pattern Recognition (CVPR)}, 2023.

\bibitem[Brahmbhatt et~al.(2019)Brahmbhatt, Handa, Hays, and Fox]{brahmbhatt2019contactgrasp}
Samarth Brahmbhatt, Ankur Handa, James Hays, and Dieter Fox.
\newblock Contactgrasp: Functional multi-finger grasp synthesis from contact.
\newblock In \emph{2019 IEEE/RSJ International Conference on Intelligent Robots and Systems (IROS)}, pages 2386--2393. IEEE, 2019.

\bibitem[Braun et~al.(2023)Braun, Christen, Kocabas, Aksan, and Hilliges]{braun2023physically}
Jona Braun, Sammy Christen, Muhammed Kocabas, Emre Aksan, and Otmar Hilliges.
\newblock Physically plausible full-body hand-object interaction synthesis.
\newblock \emph{arXiv preprint arXiv:2309.07907}, 2023.

\bibitem[Christen et~al.(2024)Christen, Hampali, Sener, Remelli, Hodan, Sauser, Ma, and Tekin]{christen2024diffh2o}
Sammy Christen, Shreyas Hampali, Fadime Sener, Edoardo Remelli, Tomas Hodan, Eric Sauser, Shugao Ma, and Bugra Tekin.
\newblock Diffh2o: Diffusion-based synthesis of hand-object interactions from textual descriptions.
\newblock \emph{arXiv preprint arXiv:2403.17827}, 2024.

\bibitem[Diller and Dai(2023)]{diller2023cg}
Christian Diller and Angela Dai.
\newblock Cg-hoi: Contact-guided 3d human-object interaction generation.
\newblock \emph{arXiv preprint arXiv:2311.16097}, 2023.

\bibitem[Ding et~al.(2023)Ding, Zhang, Paxton, and Zhang]{ding2023task}
Yan Ding, Xiaohan Zhang, Chris Paxton, and Shiqi Zhang.
\newblock Task and motion planning with large language models for object rearrangement.
\newblock In \emph{2023 IEEE/RSJ International Conference on Intelligent Robots and Systems (IROS)}, pages 2086--2092. IEEE, 2023.

\bibitem[Fan et~al.(2023)Fan, Taheri, Tzionas, Kocabas, Kaufmann, Black, and Hilliges]{fan2023arctic}
Zicong Fan, Omid Taheri, Dimitrios Tzionas, Muhammed Kocabas, Manuel Kaufmann, Michael~J. Black, and Otmar Hilliges.
\newblock {ARCTIC}: A dataset for dexterous bimanual hand-object manipulation.
\newblock In \emph{Conference on Computer Vision and Pattern Recognition (CVPR)}, 2023.

\bibitem[Ghosh et~al.(2023{\natexlab{a}})Ghosh, Dabral, Golyanik, Theobalt, and Slusallek]{ghosh2022imos}
Anindita Ghosh, Rishabh Dabral, Vladislav Golyanik, Christian Theobalt, and Philipp Slusallek.
\newblock Imos: Intent-driven full-body motion synthesis for human-object interactions.
\newblock In \emph{Eurographics}, 2023{\natexlab{a}}.

\bibitem[Ghosh et~al.(2023{\natexlab{b}})Ghosh, Dabral, Golyanik, Theobalt, and Slusallek]{ghosh2023imos}
Anindita Ghosh, Rishabh Dabral, Vladislav Golyanik, Christian Theobalt, and Philipp Slusallek.
\newblock Imos: Intent-driven full-body motion synthesis for human-object interactions.
\newblock In \emph{Computer Graphics Forum}, pages 1--12. Wiley Online Library, 2023{\natexlab{b}}.

\bibitem[Grady et~al.(2021)Grady, Tang, Twigg, Vo, Brahmbhatt, and Kemp]{grady2021contactopt}
Patrick Grady, Chengcheng Tang, Christopher~D Twigg, Minh Vo, Samarth Brahmbhatt, and Charles~C Kemp.
\newblock Contactopt: Optimizing contact to improve grasps.
\newblock In \emph{Proceedings of the IEEE/CVF Conference on Computer Vision and Pattern Recognition}, pages 1471--1481, 2021.

\bibitem[Guo et~al.(2022)Guo, Zou, Zuo, Wang, Ji, Li, and Cheng]{guo2022generating}
Chuan Guo, Shihao Zou, Xinxin Zuo, Sen Wang, Wei Ji, Xingyu Li, and Li Cheng.
\newblock Generating diverse and natural 3d human motions from text.
\newblock In \emph{Proceedings of the IEEE/CVF Conference on Computer Vision and Pattern Recognition}, pages 5152--5161, 2022.

\bibitem[Hassan et~al.(2019)Hassan, Choutas, Tzionas, and Black]{prox}
Mohamed Hassan, Vasileios Choutas, Dimitrios Tzionas, and Michael~J Black.
\newblock Resolving 3d human pose ambiguities with 3d scene constraints.
\newblock In \emph{International Conference on Computer Vision (ICCV)}, pages 2282--2292, 2019.

\bibitem[Hassan et~al.(2021)Hassan, Ceylan, Villegas, Saito, Yang, Zhou, and Black]{hassan_samp_2021}
Mohamed Hassan, Duygu Ceylan, Ruben Villegas, Jun Saito, Jimei Yang, Yi Zhou, and Michael Black.
\newblock Stochastic scene-aware motion prediction.
\newblock In \emph{International Conference on Computer Vision (ICCV)}, pages 11354--11364, 2021.

\bibitem[Hassan et~al.(2023)Hassan, Guo, Wang, Black, Fidler, and Peng]{hassan2023synthesizing}
Mohamed Hassan, Yunrong Guo, Tingwu Wang, Michael Black, Sanja Fidler, and Xue~Bin Peng.
\newblock Synthesizing physical character-scene interactions.
\newblock In \emph{SIGGRAPH 2023 Conference Papers}, 2023.

\bibitem[Hong et~al.(2023)Hong, Zhen, Chen, Zheng, Du, Chen, and Gan]{hong20233d}
Yining Hong, Haoyu Zhen, Peihao Chen, Shuhong Zheng, Yilun Du, Zhenfang Chen, and Chuang Gan.
\newblock 3d-llm: Injecting the 3d world into large language models.
\newblock \emph{Advances in Neural Information Processing Systems}, 36:\penalty0 20482--20494, 2023.

\bibitem[Hu et~al.(2023)Hu, Lin, Zhang, Yi, and Gao]{hu2023look}
Yingdong Hu, Fanqi Lin, Tong Zhang, Li Yi, and Yang Gao.
\newblock Look before you leap: Unveiling the power of gpt-4v in robotic vision-language planning.
\newblock \emph{arXiv preprint arXiv:2311.17842}, 2023.

\bibitem[Huang et~al.(2023)Huang, Wang, Li, Jia, Liu, Zhu, Liang, and Zhu]{huang2023diffusion}
Siyuan Huang, Zan Wang, Puhao Li, Baoxiong Jia, Tengyu Liu, Yixin Zhu, Wei Liang, and Song-Chun Zhu.
\newblock Diffusion-based generation, optimization, and planning in 3d scenes.
\newblock In \emph{Conference on Computer Vision and Pattern Recognition (CVPR)}, 2023.

\bibitem[Jiang et~al.(2021)Jiang, Liu, Wang, and Wang]{jiang2021hand}
Hanwen Jiang, Shaowei Liu, Jiashun Wang, and Xiaolong Wang.
\newblock Hand-object contact consistency reasoning for human grasps generation.
\newblock In \emph{Proceedings of the IEEE/CVF international conference on computer vision}, pages 11107--11116, 2021.

\bibitem[Jiang et~al.(2024{\natexlab{a}})Jiang, He, Wang, Li, Chen, Huang, and Zhu]{jiang2024autonomous}
Nan Jiang, Zimo He, Zi Wang, Hongjie Li, Yixin Chen, Siyuan Huang, and Yixin Zhu.
\newblock Autonomous character-scene interaction synthesis from text instruction.
\newblock \emph{arXiv preprint arXiv:2410.03187}, 2024{\natexlab{a}}.

\bibitem[Jiang et~al.(2024{\natexlab{b}})Jiang, Zhang, Li, Ma, Wang, Chen, Liu, Zhu, and Huang]{jiang2024scaling}
Nan Jiang, Zhiyuan Zhang, Hongjie Li, Xiaoxuan Ma, Zan Wang, Yixin Chen, Tengyu Liu, Yixin Zhu, and Siyuan Huang.
\newblock Scaling up dynamic human-scene interaction modeling.
\newblock \emph{arXiv preprint arXiv:2403.08629}, 2024{\natexlab{b}}.

\bibitem[Johnson et~al.(2015)Johnson, Krishna, Stark, Li, Shamma, Bernstein, and Fei-Fei]{johnson2015image}
Justin Johnson, Ranjay Krishna, Michael Stark, Li-Jia Li, David Shamma, Michael Bernstein, and Li Fei-Fei.
\newblock Image retrieval using scene graphs.
\newblock In \emph{Proceedings of the IEEE conference on computer vision and pattern recognition}, pages 3668--3678, 2015.

\bibitem[Karunratanakul et~al.(2020)Karunratanakul, Yang, Zhang, Black, Muandet, and Tang]{karunratanakul2020grasping}
Korrawe Karunratanakul, Jinlong Yang, Yan Zhang, Michael~J Black, Krikamol Muandet, and Siyu Tang.
\newblock Grasping field: Learning implicit representations for human grasps.
\newblock In \emph{2020 International Conference on 3D Vision (3DV)}, pages 333--344. IEEE, 2020.

\bibitem[Kingma and Ba(2015)]{kingma2014adam}
Diederik~P Kingma and Jimmy Ba.
\newblock Adam: A method for stochastic optimization.
\newblock In \emph{International Conference on Learning Representations (ICLR)}, 2015.

\bibitem[Li et~al.(2023{\natexlab{a}})Li, Clegg, Mottaghi, Wu, Puig, and Liu]{li2023controllable}
Jiaman Li, Alexander Clegg, Roozbeh Mottaghi, Jiajun Wu, Xavier Puig, and C~Karen Liu.
\newblock Controllable human-object interaction synthesis.
\newblock \emph{arXiv preprint arXiv:2312.03913}, 2023{\natexlab{a}}.

\bibitem[Li et~al.(2023{\natexlab{b}})Li, Wu, and Liu]{li2023object}
Jiaman Li, Jiajun Wu, and C~Karen Liu.
\newblock Object motion guided human motion synthesis.
\newblock \emph{ACM Trans. Graph.}, 42\penalty0 (6), 2023{\natexlab{b}}.

\bibitem[Li et~al.(2024)Li, Wang, Loy, and Dai]{li2024task}
Quanzhou Li, Jingbo Wang, Chen~Change Loy, and Bo Dai.
\newblock Task-oriented human-object interactions generation with implicit neural representations.
\newblock In \emph{Proceedings of the IEEE/CVF Winter Conference on Applications of Computer Vision}, pages 3035--3044, 2024.

\bibitem[Li et~al.(2007)Li, Fu, and Pollard]{li2007data}
Ying Li, Jiaxin~L Fu, and Nancy~S Pollard.
\newblock Data-driven grasp synthesis using shape matching and task-based pruning.
\newblock \emph{IEEE Transactions on Visualization and Computer Graphics}, 13\penalty0 (4):\penalty0 732--747, 2007.

\bibitem[Liu et~al.(2024)Liu, Li, Wu, and Lee]{liu2024visual}
Haotian Liu, Chunyuan Li, Qingyang Wu, and Yong~Jae Lee.
\newblock Visual instruction tuning.
\newblock \emph{Advances in neural information processing systems}, 36, 2024.

\bibitem[Liu and Hodgins(2018)]{liu2018learning}
Libin Liu and Jessica Hodgins.
\newblock Learning basketball dribbling skills using trajectory optimization and deep reinforcement learning.
\newblock \emph{ACM Transactions on Graphics (TOG)}, 37\penalty0 (4):\penalty0 1--14, 2018.

\bibitem[Liu and Yi(2024)]{liu2024geneoh}
Xueyi Liu and Li Yi.
\newblock Geneoh diffusion: Towards generalizable hand-object interaction denoising via denoising diffusion.
\newblock \emph{arXiv preprint arXiv:2402.14810}, 2024.

\bibitem[Luo et~al.(2024)Luo, Cao, Christen, Winkler, Kitani, and Xu]{luo2024grasping}
Zhengyi Luo, Jinkun Cao, Sammy Christen, Alexander Winkler, Kris Kitani, and Weipeng Xu.
\newblock Grasping diverse objects with simulated humanoids.
\newblock \emph{arXiv preprint arXiv:2407.11385}, 2024.

\bibitem[Makoviychuk et~al.(2021)Makoviychuk, Wawrzyniak, Guo, Lu, Storey, Macklin, Hoeller, Rudin, Allshire, Handa, and State]{makoviychuk2021isaac}
Viktor Makoviychuk, Lukasz Wawrzyniak, Yunrong Guo, Michelle Lu, Kier Storey, Miles Macklin, David Hoeller, Nikita Rudin, Arthur Allshire, Ankur Handa, and Gavriel State.
\newblock Isaac gym: High performance gpu-based physics simulation for robot learning, 2021.

\bibitem[Mir et~al.(2023)Mir, Puig, Kanazawa, and Pons-Moll]{mir2023generating}
Aymen Mir, Xavier Puig, Angjoo Kanazawa, and Gerard Pons-Moll.
\newblock Generating continual human motion in diverse 3d scenes.
\newblock \emph{arXiv preprint arXiv:2304.02061}, 2023.

\bibitem[OpenAI(2024)]{gpt4o}
OpenAI.
\newblock Hello gpt-4o, 2024.

\bibitem[OpenAI(2023)]{openai2023gpt}
R OpenAI.
\newblock Gpt-4 technical report. arxiv 2303.08774.
\newblock \emph{View in Article}, 2\penalty0 (5), 2023.

\bibitem[Pavlakos et~al.(2019)Pavlakos, Choutas, Ghorbani, Bolkart, Osman, Tzionas, and Black]{smplx}
Georgios Pavlakos, Vasileios Choutas, Nima Ghorbani, Timo Bolkart, Ahmed A.~A. Osman, Dimitrios Tzionas, and Michael~J. Black.
\newblock Expressive body capture: 3d hands, face, and body from a single image.
\newblock In \emph{Conference on Computer Vision and Pattern Recognition (CVPR)}, pages 10975--10985, 2019.

\bibitem[Peng et~al.(2023)Peng, Xie, Wu, Jampani, Sun, and Jiang]{peng2023hoi}
Xiaogang Peng, Yiming Xie, Zizhao Wu, Varun Jampani, Deqing Sun, and Huaizu Jiang.
\newblock Hoi-diff: Text-driven synthesis of 3d human-object interactions using diffusion models.
\newblock \emph{arXiv preprint arXiv:2312.06553}, 2023.

\bibitem[Peng et~al.(2018)Peng, Abbeel, Levine, and Van~de Panne]{peng2018deepmimic}
Xue~Bin Peng, Pieter Abbeel, Sergey Levine, and Michiel Van~de Panne.
\newblock Deepmimic: Example-guided deep reinforcement learning of physics-based character skills.
\newblock \emph{ACM Transactions On Graphics (TOG)}, 37\penalty0 (4):\penalty0 1--14, 2018.

\bibitem[Peng et~al.(2021)Peng, Ma, Abbeel, Levine, and Kanazawa]{peng2021amp}
Xue~Bin Peng, Ze Ma, Pieter Abbeel, Sergey Levine, and Angjoo Kanazawa.
\newblock Amp: Adversarial motion priors for stylized physics-based character control.
\newblock \emph{ACM Transactions on Graphics (TOG)}, 40\penalty0 (4):\penalty0 1--20, 2021.

\bibitem[Pollard and Zordan(2005)]{pollard2005physically}
Nancy~S Pollard and Victor~Brian Zordan.
\newblock Physically based grasping control from example.
\newblock In \emph{Proceedings of the 2005 ACM SIGGRAPH/Eurographics symposium on Computer animation}, pages 311--318, 2005.

\bibitem[Prokudin et~al.(2019)Prokudin, Lassner, and Romero]{prokudin2019efficient}
Sergey Prokudin, Christoph Lassner, and Javier Romero.
\newblock Efficient learning on point clouds with basis point sets.
\newblock In \emph{International Conference on Computer Vision (ICCV)}, pages 4332--4341, 2019.

\bibitem[Radford et~al.(2021)Radford, Kim, Hallacy, Ramesh, Goh, Agarwal, Sastry, Askell, Mishkin, Clark, et~al.]{radford2021learning}
Alec Radford, Jong~Wook Kim, Chris Hallacy, Aditya Ramesh, Gabriel Goh, Sandhini Agarwal, Girish Sastry, Amanda Askell, Pamela Mishkin, Jack Clark, et~al.
\newblock Learning transferable visual models from natural language supervision.
\newblock In \emph{International conference on machine learning}, pages 8748--8763. PMLR, 2021.

\bibitem[Rodriguez et~al.(2012)Rodriguez, Mason, and Ferry]{rodriguez2012caging}
Alberto Rodriguez, Matthew~T Mason, and Steve Ferry.
\newblock From caging to grasping.
\newblock \emph{The International Journal of Robotics Research}, 31\penalty0 (7):\penalty0 886--900, 2012.

\bibitem[Schulman et~al.(2017)Schulman, Wolski, Dhariwal, Radford, and Klimov]{schulman2017proximal}
John Schulman, Filip Wolski, Prafulla Dhariwal, Alec Radford, and Oleg Klimov.
\newblock Proximal policy optimization algorithms, 2017.

\bibitem[Taheri et~al.(2020)Taheri, Ghorbani, Black, and Tzionas]{GRAB:2020}
Omid Taheri, Nima Ghorbani, Michael~J. Black, and Dimitrios Tzionas.
\newblock {GRAB}: A dataset of whole-body human grasping of objects.
\newblock In \emph{European Conference on Computer Vision (ECCV)}, 2020.

\bibitem[Taheri et~al.(2022)Taheri, Choutas, Black, and Tzionas]{taheri2022goal}
Omid Taheri, Vasileios Choutas, Michael~J Black, and Dimitrios Tzionas.
\newblock Goal: Generating 4d whole-body motion for hand-object grasping.
\newblock In \emph{Conference on Computer Vision and Pattern Recognition (CVPR)}, pages 13263--13273, 2022.

\bibitem[Taheri et~al.(2023)Taheri, Zhou, Tzionas, Zhou, Ceylan, Pirk, and Black]{taheri2023grip}
Omid Taheri, Yi Zhou, Dimitrios Tzionas, Yang Zhou, Duygu Ceylan, Soren Pirk, and Michael~J Black.
\newblock Grip: Generating interaction poses using latent consistency and spatial cues.
\newblock \emph{arXiv preprint arXiv:2308.11617}, 2023.

\bibitem[Tendulkar et~al.(2023)Tendulkar, Sur{\'\i}s, and Vondrick]{tendulkar2023flex}
Purva Tendulkar, D{\'\i}dac Sur{\'\i}s, and Carl Vondrick.
\newblock Flex: Full-body grasping without full-body grasps.
\newblock In \emph{Proceedings of the IEEE/CVF Conference on Computer Vision and Pattern Recognition}, pages 21179--21189, 2023.

\bibitem[Wake et~al.(2023)Wake, Kanehira, Sasabuchi, Takamatsu, and Ikeuchi]{wake2023gpt}
Naoki Wake, Atsushi Kanehira, Kazuhiro Sasabuchi, Jun Takamatsu, and Katsushi Ikeuchi.
\newblock Gpt-4v (ision) for robotics: Multimodal task planning from human demonstration.
\newblock \emph{arXiv preprint arXiv:2311.12015}, 2023.

\bibitem[Wang et~al.(2021)Wang, Xu, Xu, Liu, and Wang]{wang2021synthesizing}
Jiashun Wang, Huazhe Xu, Jingwei Xu, Sifei Liu, and Xiaolong Wang.
\newblock Synthesizing long-term 3d human motion and interaction in 3d scenes.
\newblock In \emph{Conference on Computer Vision and Pattern Recognition (CVPR)}, pages 9401--9411, 2021.

\bibitem[Wang et~al.(2022{\natexlab{a}})Wang, Rong, Liu, Yan, Lin, and Dai]{wang2022towards}
Jingbo Wang, Yu Rong, Jingyuan Liu, Sijie Yan, Dahua Lin, and Bo Dai.
\newblock Towards diverse and natural scene-aware 3d human motion synthesis.
\newblock In \emph{Conference on Computer Vision and Pattern Recognition (CVPR)}, pages 20460--20469, 2022{\natexlab{a}}.

\bibitem[Wang et~al.(2023{\natexlab{a}})Wang, Zhang, Chen, Xu, Li, Liu, and Wang]{wang2023dexgraspnet}
Ruicheng Wang, Jialiang Zhang, Jiayi Chen, Yinzhen Xu, Puhao Li, Tengyu Liu, and He Wang.
\newblock Dexgraspnet: A large-scale robotic dexterous grasp dataset for general objects based on simulation.
\newblock In \emph{2023 IEEE International Conference on Robotics and Automation (ICRA)}, pages 11359--11366. IEEE, 2023{\natexlab{a}}.

\bibitem[Wang et~al.(2024{\natexlab{a}})Wang, Xu, Shi, Schumann, and Liu]{wang2024f}
Ruocheng Wang, Pei Xu, Haochen Shi, Elizabeth Schumann, and C~Karen Liu.
\newblock F$\backslash$" urelise: Capturing and physically synthesizing hand motions of piano performance.
\newblock \emph{arXiv preprint arXiv:2410.05791}, 2024{\natexlab{a}}.

\bibitem[Wang et~al.(2023{\natexlab{b}})Wang, Lin, Zeng, Luo, Zhang, and Zhang]{wang2023physhoi}
Yinhuai Wang, Jing Lin, Ailing Zeng, Zhengyi Luo, Jian Zhang, and Lei Zhang.
\newblock Physhoi: Physics-based imitation of dynamic human-object interaction.
\newblock \emph{arXiv preprint arXiv:2312.04393}, 2023{\natexlab{b}}.

\bibitem[Wang et~al.(2022{\natexlab{b}})Wang, Chen, Liu, Zhu, Liang, and Huang]{wang2022humanise}
Zan Wang, Yixin Chen, Tengyu Liu, Yixin Zhu, Wei Liang, and Siyuan Huang.
\newblock Humanise: Language-conditioned human motion generation in 3d scenes.
\newblock In \emph{Advances in Neural Information Processing Systems (NeurIPS)}, 2022{\natexlab{b}}.

\bibitem[Wang et~al.(2024{\natexlab{b}})Wang, Chen, Jia, Li, Zhang, Zhang, Liu, Zhu, Liang, and Huang]{wang2024move}
Zan Wang, Yixin Chen, Baoxiong Jia, Puhao Li, Jinlu Zhang, Jingze Zhang, Tengyu Liu, Yixin Zhu, Wei Liang, and Siyuan Huang.
\newblock Move as you say, interact as you can: Language-guided human motion generation with scene affordance.
\newblock \emph{arXiv preprint arXiv:2403.18036}, 2024{\natexlab{b}}.

\bibitem[Wu et~al.(2024)Wu, Shi, Huang, Yu, Xu, and Wang]{wu2024thor}
Qianyang Wu, Ye Shi, Xiaoshui Huang, Jingyi Yu, Lan Xu, and Jingya Wang.
\newblock Thor: Text to human-object interaction diffusion via relation intervention.
\newblock \emph{arXiv preprint arXiv:2403.11208}, 2024.

\bibitem[Wu et~al.(2022)Wu, Wang, Zhang, Zhang, Hilliges, Yu, and Tang]{wu2022saga}
Yan Wu, Jiahao Wang, Yan Zhang, Siwei Zhang, Otmar Hilliges, Fisher Yu, and Siyu Tang.
\newblock Saga: Stochastic whole-body grasping with contact.
\newblock In \emph{European Conference on Computer Vision (ECCV)}, pages 257--274, 2022.

\bibitem[Xiao et~al.(2024)Xiao, Wang, Wang, Cao, Zhang, Dai, Lin, and Pang]{xiao2024unified}
Zeqi Xiao, Tai Wang, Jingbo Wang, Jinkun Cao, Wenwei Zhang, Bo Dai, Dahua Lin, and Jiangmiao Pang.
\newblock Unified human-scene interaction via prompted chain-of-contacts.
\newblock In \emph{The Twelfth International Conference on Learning Representations}, 2024.

\bibitem[Xie et~al.(2022)Xie, Starke, Ling, and van~de Panne]{xie2022learning}
Zhaoming Xie, Sebastian Starke, Hung~Yu Ling, and Michiel van~de Panne.
\newblock Learning soccer juggling skills with layer-wise mixture-of-experts.
\newblock In \emph{ACM SIGGRAPH 2022 Conference Proceedings}, pages 1--9, 2022.

\bibitem[Xu and Karamouzas(2021)]{xu2021gan}
Pei Xu and Ioannis Karamouzas.
\newblock A gan-like approach for physics-based imitation learning and interactive character control.
\newblock \emph{Proceedings of the ACM on Computer Graphics and Interactive Techniques}, 4\penalty0 (3):\penalty0 1--22, 2021.

\bibitem[Xu and Wang(2024)]{xu2024synchronize}
Pei Xu and Ruocheng Wang.
\newblock Synchronize dual hands for physics-based dexterous guitar playing.
\newblock \emph{arXiv preprint arXiv:2409.16629}, 2024.

\bibitem[Xu et~al.(2024)Xu, Wang, Wang, and Gui]{xu2024interdreamer}
Sirui Xu, Ziyin Wang, Yu-Xiong Wang, and Liang-Yan Gui.
\newblock Interdreamer: Zero-shot text to 3d dynamic human-object interaction.
\newblock \emph{arXiv preprint arXiv:2403.19652}, 2024.

\bibitem[Yang et~al.(2024)Yang, Sun, Weihs, VanderBilt, Herrasti, Han, Wu, Haber, Krishna, Liu, et~al.]{yang2024holodeck}
Yue Yang, Fan-Yun Sun, Luca Weihs, Eli VanderBilt, Alvaro Herrasti, Winson Han, Jiajun Wu, Nick Haber, Ranjay Krishna, Lingjie Liu, et~al.
\newblock Holodeck: Language guided generation of 3d embodied ai environments.
\newblock In \emph{Proceedings of the IEEE/CVF Conference on Computer Vision and Pattern Recognition}, pages 16227--16237, 2024.

\bibitem[Yao et~al.(2024)Yao, Song, Zhou, Ao, Chen, and Liu]{yao2024moconvq}
Heyuan Yao, Zhenhua Song, Yuyang Zhou, Tenglong Ao, Baoquan Chen, and Libin Liu.
\newblock Moconvq: Unified physics-based motion control via scalable discrete representations.
\newblock \emph{ACM Transactions on Graphics (TOG)}, 43\penalty0 (4):\penalty0 1--21, 2024.

\bibitem[Ye and Liu(2012)]{ye2012synthesis}
Yuting Ye and C~Karen Liu.
\newblock Synthesis of detailed hand manipulations using contact sampling.
\newblock \emph{ACM Transactions on Graphics (ToG)}, 31\penalty0 (4):\penalty0 1--10, 2012.

\bibitem[Yi et~al.(2024)Yi, Thies, Black, Peng, and Rempe]{yi2024generating}
Hongwei Yi, Justus Thies, Michael~J Black, Xue~Bin Peng, and Davis Rempe.
\newblock Generating human interaction motions in scenes with text control.
\newblock \emph{arXiv preprint arXiv:2404.10685}, 2024.

\bibitem[Zeng et~al.(2023)Zeng, Wu, Yang, Zhang, Ding, Cheng, and Dong]{zeng2023distilling}
Yiming Zeng, Mingdong Wu, Long Yang, Jiyao Zhang, Hao Ding, Hui Cheng, and Hao Dong.
\newblock Distilling functional rearrangement priors from large models.
\newblock \emph{arXiv preprint arXiv:2312.01474}, 2023.

\bibitem[Zhang et~al.(2021)Zhang, Ye, Shiratori, and Komura]{zhang2021manipnet}
He Zhang, Yuting Ye, Takaaki Shiratori, and Taku Komura.
\newblock Manipnet: neural manipulation synthesis with a hand-object spatial representation.
\newblock \emph{ACM Transactions on Graphics (ToG)}, 40\penalty0 (4):\penalty0 1--14, 2021.

\bibitem[Zhang et~al.(2022)Zhang, Bhatnagar, Starke, Guzov, and Pons-Moll]{zhang2022couch}
Xiaohan Zhang, Bharat~Lal Bhatnagar, Sebastian Starke, Vladimir Guzov, and Gerard Pons-Moll.
\newblock Couch: Towards controllable human-chair interactions.
\newblock In \emph{European Conference on Computer Vision (ECCV)}, pages 518--535, 2022.

\bibitem[Zhang et~al.(2024)Zhang, Huang, Liu, Tang, Lu, Chen, Bai, Chu, Yu, and Ouyang]{zhang2024motiongpt}
Yaqi Zhang, Di Huang, Bin Liu, Shixiang Tang, Yan Lu, Lu Chen, Lei Bai, Qi Chu, Nenghai Yu, and Wanli Ouyang.
\newblock Motiongpt: Finetuned llms are general-purpose motion generators.
\newblock In \emph{Proceedings of the AAAI Conference on Artificial Intelligence}, pages 7368--7376, 2024.

\bibitem[Zhao et~al.(2023)Zhao, Zhang, Wang, Beeler, and Tang]{zhao2023synthesizing}
Kaifeng Zhao, Yan Zhang, Shaofei Wang, Thabo Beeler, and Siyu Tang.
\newblock Synthesizing diverse human motions in 3d indoor scenes.
\newblock \emph{arXiv preprint arXiv:2305.12411}, 2023.

\bibitem[Zhou et~al.(2019)Zhou, Barnes, Lu, Yang, and Li]{zhou2019continuity}
Yi Zhou, Connelly Barnes, Jingwan Lu, Jimei Yang, and Hao Li.
\newblock On the continuity of rotation representations in neural networks.
\newblock In \emph{Computer Vision and Pattern Recognition (CVPR)}, 2019.

\end{thebibliography}
